\documentclass[sn-apa]{sn-jnl}

\usepackage{caption}
\usepackage{subcaption}
\usepackage{xcolor}

\jyear{2022}%

\theoremstyle{thmstyleone}%
\theoremstyle{thmstyletwo}%

\theoremstyle{thmstylethree}%

\usepackage{natbib}
\setcitestyle{authoryear, round, citesep={,},aysep={}}

\usepackage{enumitem}

\begin{document}

\title{Reformulating van Rijsbergen's $F_\beta$ metric for weighted binary cross-entropy}


\author{\fnm{Satesh Ramdhani} }\email{satesh.ramdhani@gmail.com}




\abstract{The separation of performance metrics from gradient based loss functions may not always give optimal results and may miss vital aggregate information. This paper investigates incorporating a performance metric alongside differentiable loss functions to inform training outcomes. The goal is to guide model performance and interpretation by assuming statistical distributions on this performance metric for dynamic weighting. The focus is on van Rijsbergen’s $\text{F}_\beta$ metric -- a popular choice for gauging classification performance. Through distributional assumptions on the $\text{F}_\beta$, an intermediary link can be established to the standard binary cross-entropy via dynamic penalty weights. First, the $\text{F}_\beta$ metric is reformulated to facilitate assuming statistical distributions with accompanying proofs for the cumulative density function. These probabilities are used within a knee curve algorithm to find an optimal $\beta$ or $\beta_{opt}$. This $\beta_{opt}$ is used as a weight or penalty in the proposed weighted binary cross-entropy. Experimentation on publicly available data along with benchmark analysis mostly yields better and interpretable results as compared to the baseline for both imbalanced and balanced classes. For example, for the IMDB text data with known labeling errors, a 14\% boost in $F_1$ score is shown. The results also reveal commonalities between the penalty model families derived in this paper and the suitability of recall-centric or precision-centric parameters used in the optimization. The flexibility of this methodology can enhance interpretation.}

\keywords{Performance metrics, Metrics, F-Beta Metric, Penalty Optimization, C.J. van Rijsbergen, Information Retrieval, Weighted Cross-Entropy, Binary Cross-Entropy, Text Retrieval}

\maketitle
\newpage

\section{Acronym List}
    
    \begin{description}[leftmargin=*, widest=DCCHTM]
    
        \item[$F_\beta$]
        F-Beta Metric
        
        \item[$\beta_{opt}$]
        Optimal $\beta$ from Algorithm \ref{algo1}

        \item[$M_1^\beta$]
        Model 1: U \& IU from (\ref{sub1sec4})
        
        \item[$M_2^{(\lambda, \sigma^2)}$]
        Model 2: Ga \& IE from (\ref{sub2sec4})
         
        \item[PV]
        Pressure Vessel Design 
        
        \item[$T_s$]
        Thickness of the pressure vessel shell
        
        \item[$T_h$]
        Thickness of the pressure vessel head 
        
        \item[UST]
        Underground Storage Tank

        \item[$f_{C}$]
        Equation for the volume a cylindrical UST (\ref{fc})
        
        \item[$f_{CH}$]
        Equation for the volume of a cylindrical UST with hemispherical endcaps (\ref{fch})
        
        \item[$f_{ED}$]
        Equation for the volume of an ellipsoidal UST (\ref{fed})
        
        \item[$f_{EDH}$]
        Equation for the volume of an ellipsoidal UST with hemi-ellipsoidal end-caps (\ref{fedh})
      
        \item[CvE]
        Cylindrical UST versus Ellipsoidal UST
        
        \item[CHvEH]
        Cylindrical UST with hemispherical endcaps versus ellipsoidal UST with hemi-ellipsoidal end-caps
        
         \item[UCI]
        UCI Machine Learning Repository
    \end{description}
  
\section{Introduction}\label{sec1} 
Data imbalance is a known, and widespread real world issue that affects performance metrics for a variety of learning algorithm problems (i.e., image detection and segmentation, text categorization and classification). Approaches to mitigate this issue generally fall into three categories: adjusting the neural network architecture (including multiple models or ensembles like \citealt{f1_3}), adjusting the loss function used for training, or adjusting the data (i.e., collecting more data, or leveraging sampling techniques like \citealt{smote} and \citealt{overundersample}). This research looks at adjusting the loss function with a focus on incorporating the $\text{F}_\beta$ performance metric. The interconnection between performance metric and loss function is crucial for understanding both model behavior and the inherent nature of that specific dataset. This connection has already been approached from the angle of thresholding (a post model step) as in \citet{threshold} or developing a problem specific metric, as \citet{metriclikelrp}, \citet{dice}, and \citet{lrp} did for real world mislabeling costs, dynamic weighting for easy negative samples, and object detection, respectively. This paper takes a uniquely different and novel approach where statistical distributions act as an intermediary to connect the $\text{F}_\beta$ metric to the binary cross-entropy through dynamic penalty weights. 

First, the derivation of the $F_\beta$ metric from van Rijsbergen’s effectiveness score, $E$, is revisited to prove a limiting case of $F_1$ in section \ref{sec3}. This result supports the default case for the main algorithm in section \ref{sec5}. 

Second, the $F_\beta$ metric is reformulated into a multiplicative form by assuming two independent random variables. Then parametric statistical distributions are assumed for these random variables. In particular, the Uniform and Inverse Uniform (U \& IU) case and the Gaussian and Inverse Exponential (Ga \& IE) case are proposed. The idea behind U \& IU is that no known insight is assumed on the $F_\beta$ cumulative density function's (CDF) surface. But the Ga \& IE provides the practitioner more flexibility in setting some insight to this CDF surface. This leads to a more interpretable performance metric that is configurable to the data without having to create a new problem specific metric (or loss function).

Third, for both distributional cases, the CDF or $\text{Pr}\left( F_\beta \right)$ shown in section \ref{sec4} facilitates finding an optimal $\beta$ through a knee curve algorithm in section \ref{sub1sec5}. This algorithm gets the best $\beta$ from a monotonic knee curve given precision and recall. It is the value when the curve levels off. The $\beta_{opt}$ surface for different parameter settings found in section \ref{sub3sec5} suggests a slightly more recall centric penalty. This is discussed further in section \ref{sec6}.

Finally, a weighted binary cross-entropy loss function based on $\beta_{opt}$ is proposed in section \ref{sub2sec5}. This loss methodology is applied to three data categories: image, text and tabular/structured data. For contextual data (i.e., image and text), model performance for $F_1$ improves, and the best result occurs for the text data that contains (known) labeling errors. The structured/tabular or non-contextual data does not show significant $F_1$ improvement, but provides an important result: when considering neural embedding architectures for training, the type (or category) of data matters.

\section{Related Work}\label{sec2}
Logistic regression models are one of the most fundamental statistically based classifier. \cite{f1_2} provides a training procedure that uses a sigmoid approximation to maximize the  $F_\beta$ on this class of classifiers. When comparing the surface plots of the likelihood from \citeauthor{f1_2} and that from section \ref{sec4} -- a similar but not an equivalent comparison -- a comparable rate of change can be seen for both surfaces with respect to their respective parameters. This is an important similarity because this paper's procedure applies distributional assumptions to provide dynamic penalties to a well-known binary cross-entropy loss. Also, implementation of this paper's methodology is straightforward because it avoids the need to provide updated partial derivatives for the loss function. Furthermore, \citeauthor{f1_2} alludes to (future work that considers) a general method to optimize several operating points simultaneously, which is a fundamental and indirect assertion in this paper. The sigmoid approximation is also used by \cite{f1_3} in the multi-label setting for text categorization. In their framework, multiple binary classifiers are trained per category and combined with weights estimated to maximize micro- or macro-averaged $F_1$ scores.

Similarly, \cite{bib102} propose a methodology for performance metric learning that uses a metric approximation (i.e., AUC, $F_1$) derived from the confusion matrix. The back-propagation error term involves the first derivative, followed by the application of gradient descent. This method provides an alternative means of integrating performance metrics with gradient-based learning. However, there are cases where the back-propagation term proposed by \citeauthor{bib102} may pose issues. For instance, when considering equation 13 from \citeauthor{bib102} in conjunction with batch training and severe imbalance, there could be a division by zero error if a batch with only the zero label appears. Moreover, \citeauthor{bib102} test several metrics: $F_1$, G-mean, AG-mean and AUC for their method. But the G-mean, AG-mean and AUC, based on the confusion matrix approximation, can be derived as functions of $F_1$. This suggests that $F_\beta$ is more flexible than G-mean, AG-mean and AUC. In other words, $\beta$ is unique to $F_\beta$ yet generalized across other metrics when equal to 1. In fact, for class imbalance, the AUC metric - an average over many thresholds, and G-mean - a geometric mean, is less stringent and more generous in accuracy reporting compared to the $F_1$.  This is the reason all results in this paper are reported using the $F_1$ score.

Surrogate loss functions attempt to mimic certain aspects of the $F_\beta$ and is another related area. For example, \textit{sigmoidF1} from \cite{smoothF1} creates smooth versions for the entries of the confusion matrix, which is used to create a differentiable loss function that imitates the $F_1$. This smooth differentiability is another application of a sigmoid approximation similar to \citeauthor{f1_2}. \cite{bib3} formulates a surrogate loss by adjusting the cross-entropy loss such that its gradient matches the gradient of a smooth version of the  $F_\beta$. 

In terms of metric creation or variation to the  $F_\beta$, \cite{metriclikelrp}, \cite{dice}, \cite{rankandlrp} and \cite{f1_1} are highlighted. The Real World Weight Cross Entropy (RWWCE) loss function from \citeauthor{metriclikelrp} is a metric similar in spirit to \citeauthor{lrp} The idea is to set (not train or tune) cost related weights based on the dataset and the main problem, by introducing costs (i.e., financial costs) that reflect the real world. RWWCE affects both the positive and negative labels by tying each to its own real world cost implication. The dice loss from \citeauthor{dice} propose a dynamic weight adjustment to address the dominating effect of easy-negative examples. The formulation is based on the $F_\beta$ using a smoothing parameter and a focal adaptation from \cite{focal}. A ranking loss based on the Localisation Recall Precision (LRP) metric \cite{lrp} is developed by \cite{rankandlrp} for object detection. They propose an averaged LRP alongside a ranking loss function for not only classification but also localisation of objects in images. This provides a balance between both positive and negative samples. Along a similar theme, \cite{f1_1} explores a discriminative loss function that aims to maximize the expected $F_\beta$ directly for speech mispronunciation. Their loss function is based on the $F_\beta$ (comparing human assessors and the model prediction) weighted by a probability distribution (i.e., normal distribution) for that score. The final objective function is a weighted average between their loss function and the ordinary cross-entropy.

When considering the components of performance metrics, precision and recall are often the primary focus. \cite{recall2} and \cite{recall} propose two different loss functions that are both recall oriented. \citeauthor{recall2} adjust the hinge loss by adding a recall based cost (and penalty) into the objective function. As they said, by favoring recall over precision it results in a substantial boost to recall and $F_1$. By leveraging the concept of inverse frequency weighting (i.e., a sampling based technique), \citeauthor{recall} adjust the cross-entropy to reflect an inverse weighting on false negatives per class. They state that their loss function sits between regular and inverse frequency weighted cross-entropy by balancing the excessive false positives introduced by constantly up-weighting minority classes. When they consider a similar loss function using precision, this loss function shows irregular behavior. These findings are insightful because this paper's $\beta_{opt}$ surface as seen in section \ref{sub3sec5} is more recall centric with the added benefit of being able to incorporate precision weighting through the assumed probability surface.  

\section{Background}\label{sec3}
The $F_\beta$ measure comes directly from van Rijsbergen’s effectiveness score, $E$, for information retrieval (chapter 7 in \citealt{bib1}). For the theory on the six conditions supporting $E$ as a measure, refer to \citeauthor{bib1}. This paper highlights two of these conditions. First, $E$ guides the practitioner's ability to quantify effectiveness given any point ($r$, $p$) -- where $r$ and $p$ are recall and precision -- as compared to some other point. Second, precision and recall contribute effects \textit{independently} of $E$. As said by \citeauthor{bib1}, for a constant $r$ (or $p$) the difference in $E$ from any set of varying points of $p$ (or $r$) can not be removed by changing the constant. These conditions suggest equivalence relations and imply a common effectiveness (CE) curve based on precision and recall (definition 3 in \citealt{bib1}). They also motivate the rationale on using statistical distributions to understand the CE curve. The van Rijsbergen’s effectiveness measure is given in (\ref{eq1}).
\begin{equation}
E = 1 - \dfrac{1}{\alpha \frac{1}{p} + (1-\alpha) \frac{1}{r}} \label{eq1}
\end{equation}
where, $\alpha = \frac{1}{\beta^2 + 1}$. \cite{bib2} gives the details on deriving $F_\beta = \frac{(\beta^2+1)pr}{\beta^2p +r}$ from (\ref{eq1}) with $\beta = \frac{r}{p}$ and by solving $\frac{\partial E}{\partial r}$=$\frac{\partial E}{\partial p}$. The $\beta$ parameter is intended to allow the practitioner control by giving $\beta$ times more importance to recall than precision. Using the derivation steps from  \citeauthor{bib2}, a general form of $F_\beta$ for any derivative can be shown as (\ref{eq2}),
\begin{equation}
F_\beta^n (p, r) = \frac{(\beta^\frac{-2}{n-2}+1)pr}{\beta^\frac{-2}{n-2} p+r}, \label{eq2}
\end{equation}
where $n$ pertains to $\frac{\partial^n E}{\partial r^n}$=$\frac{\partial^n E}{\partial p^n}$ resulting in $\alpha_n = \frac{1}{\beta^\frac{-2}{n-2} + 1}$. Note that $n \gt 0$, and $n \neq 2$. The proof is found in Appendix \ref{secA1}. For $n=2$ the equation reduces to the equality $p=r$ implying $\beta=1$. Using (\ref{eq2}), it can be seen that the $\displaystyle{\lim_{n \to \infty} F_\beta^n = F_1^1 = \frac{2pr}{p+r}}$, which is most commonly used in the literature. The reason for showing this limiting case is to provide a justification on fixing $\beta=1$ (instead of claiming equal importance for $r$ and $p$) in the default case of any algorithm -- in particular the algorithm in section \ref{sec5}. 

\section{Reformulating the F-Beta to leverage statistical distributions}\label{sec4}
CE for neural networks is seen when different network weights give different precision and recall yet resulting in similar performance scores. CE also provides a basis for this paper's use of $\beta$ from the $F_\beta$ measure to guide training through penalties, in lieu of an explicit loss (or surrogate loss) function. In fact, \cite{bib105} uses the $F$-score as part of a preprocessing step for feature selection prior to their ensemble model (EM-PCA then ELM) for fault diagnosis. They show significant performance improvement in their approach which adds supporting evidence to this paper's use of the $F_\beta$ as a loss penalty for feature selection via gradient based learning. \\

The first step is to reformulate (\ref{eq2}) for $n=1$. This makes assuming statistical distributions easier. Consider the following reformulation through multiplicative decomposition in (\ref{eq3}) which assumes $X_1$ and $X_2$ to be independent random variables.
\begin{equation}
F_\beta = X_1 X_2, \label{eq3}
\end{equation}
where $X_1 = r^\prime + \beta^\prime$, $X_2 = (\beta^{\prime\prime} + r)^{-1}$ with $r^\prime = pr$, $\beta^\prime = \beta^2pr$ and $\beta^{\prime\prime}=\beta^2p$. $X_1$ indirectly captures imbalance in the model prediction from the underlying data. If precision and recall are on opposite ends of the $[0,1]$ scale, then $X_1$ will reflect this, while maintaining continuity when precision and recall are directionally consistent. $X_2$ can be thought of as a weighting scheme that appears recall centric with a precision based penalty. For instance, for both high (or both low) precision and recall, the weighting is consistent with intuition. However, when precision and recall are on opposite ends of the $[0,1]$ scale, the weighting sways by the aggregate with the lower score. Two use cases are considered for (\ref{eq3}): $X_1$ and $X_2$ follow U \& IU, respectively, and $X_1$ and $X_2$ follow Ga \& IE, respectively.
\subsection{Case 1: Uniform and Inverse Uniform}\label{sub1sec4}
The thought behind U \& IU is to apply (flat) equal distribution for both $X_1$ and $X_2$. These assumed distributions are applied to $\beta^\prime$ and $\beta^{\prime\prime}$ as follows:
\begin{align}
\text{Let } & \beta^\prime \sim U(0, \beta^*) \nonumber \\
                & \beta^{\prime\prime} \sim U(0, \beta^*) \nonumber \\
\text{then } & X_1 \sim U \left(r^\prime, r^\prime + \beta^* \right) \label{eq4}\\
\text{and }  & X_2 \sim IU \left(\frac{1}{r + \beta^*}, \frac{1}{r} \right), \label{eq5}
\end{align}
where $r \text{ and } r^\prime \in [0,1]$ and $\beta^* \gt 0$. Note that for both distributions there is only one $\beta^*$ chosen and this value replaces the need to have an explicit form that includes $\beta$ as a parameter. This is for convenience, as well as noticing that both $\beta^\prime$ and $\beta^{\prime\prime}$ differ by a factor of $r$. So allowing $\beta^*$ to vary broadly (which is the $\beta_{max}$ in section \ref{sec5}) would be enough to balance this convenience tradeoff. Next is to derive the joint distribution which would be used in section \ref{sec5}. It can be shown (the proof is in Appendix \ref{secA2}) that the joint distribution is:

\begin{align}
&\text{Pr}\left( F_\beta \right) = \text{Pr}\left( F_\beta \leq z \right) = 1_{\{z \leq p \& \frac{(r+\beta^*)z}{r^\prime + \beta^*} \leq 1\}} \times \dfrac{ \frac{z}{2} \left(r + \beta^* - \frac{r^\prime}{z} \right)^2}{(\beta^*)^2} \nonumber \\
&+ 1_{\{z \gt p \& \frac{(r+\beta^*)z}{r^\prime + \beta^*} \gt 1\}}  \times \left( \dfrac{rz - r^\prime}{\beta^*} + \dfrac{1}{\left( \beta^*\right)^2} \left(    \left(  r+ \beta^* + \frac{r^\prime + \beta^*}{2z}  \right) \left( r^\prime + \beta^* - rz \right)   \right. \right. \nonumber \\
&+ \left. \left. \dfrac{r \left( rz - \left( r^\prime + \beta^* \right) \right)}{2}  \right) \right) \nonumber \\
&+ 1_{\{ z \gt p \& \frac{(r+\beta^*)z}{r^\prime + \beta^*} \leq 1  \}} \times \left[ \left( \dfrac{rz - r^\prime}{\beta^*} + \dfrac{1}{\left( \beta^*\right)^2} \left(    \left(  r+ \beta^* + \frac{r^\prime + \beta^*}{2z}  \right) \left( r^\prime + \beta^* - rz \right)   \right. \right. \right. \nonumber \\
&+ \left. \left. \left. \dfrac{r \left( rz - \left( r^\prime + \beta^* \right) \right)}{2}  \right) \right) -  \dfrac{1}{\left( \beta^*\right)^2} \left( (r+\beta^*)(r^\prime +\beta^*) - \frac{(r^\prime + \beta^*)^2}{2z} - \frac{(r+\beta^*)^2 z}{2} \right) \right] \nonumber \\ \label{eq6a}
\end{align}
\begin{figure}
\centering
\begin{subfigure}[b]{0.45\textwidth}
         \centering
         \includegraphics[scale=0.15]{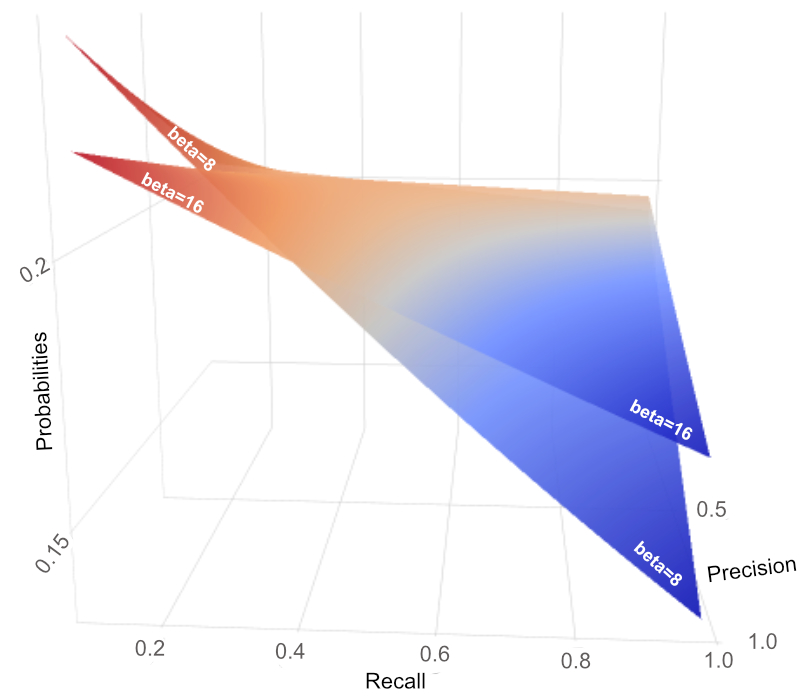}
         \caption{$P(F_\beta < 0.4)$}
         \label{fig1a}
\end{subfigure}
\begin{subfigure}[b]{0.45\textwidth}
         \centering
         \includegraphics[scale=0.15]{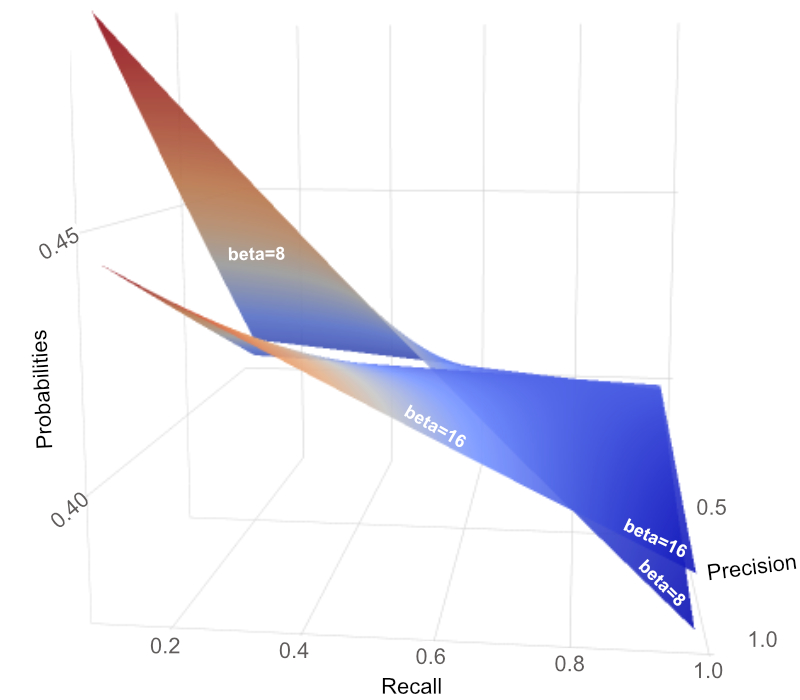}
         \caption{$P(F_\beta < 0.8)$}
        \label{fig1b}
\end{subfigure}
\caption{Probability Mass Surface: U \& IU for precision versus recall, and $\beta^* \in [8, 16]$. The cumulative probability is computed for low ($0.4$) and high ($0.8$) values.} \label{fig1}
\end{figure}
To understand this flat mixture, consider Figure \ref{fig1} - the CDF surface for a grid of precision and recall where $\beta^* \in [8, 16]$. (Note: the blue and red heat coloring is from the CDF and highlights curvature and/or rate of change). For a lower z value of $0.4$, Figure \ref{fig1a} shows that $\beta^*=8$ has a faster rate of change as compared to $\beta^*=16$. The same conclusion is apparent in Figure \ref{fig1b}, which is for a higher z value of $0.8$. For both figures more curvature is seen for lower $\beta^*$ values. This suggests that a larger $\beta^*$ value smooths the surface and is a better candidate for $\beta_{max}$ in the algorithm in Section \ref{sec5}. 

\subsection{Case 2: Gaussian and Inverse Exponential}\label{sub2sec4}
A more informed distributional approach for $X_1$ and $X_2$ considers Ga \& IE, respectively. The reason to use Gaussian distribution for $X_1$ is to allow a bell-shaped variability around a fixed $r^\prime$ that is based on $\beta^\prime$ and ultimately $\beta$. The weighting of $X_1$ by $X_2$ uses the Inverse Exponential distribution because with selections of the rate parameter $\lambda$ the distribution can shift mass from left to right as well as appear uniformly distributed around $r$. This provides practitioners enough flexibility on experimenting with different weights. The following shows the assumptions for $\beta^\prime$ and $\beta^{\prime\prime}$:
\begin{align}
\text{Let } & \beta^\prime \sim \text{Ga}(0, \sigma^2) \nonumber \\
                & \beta^{\prime\prime} \sim \text{Exponential}(\lambda) \nonumber \\
\text{then } & X_1 \sim \text{Ga}(r^\prime, \sigma^2)  \label{eq6}\\
\text{and }  & X_2 \sim \text{IE}(\lambda; r), \label{eq7}
\end{align}
where $r$ in (\ref{eq7}) is the location shift by recall from the definition of $X_2$ in (\ref{eq3}) and $\sigma^2$ is the variability captured by $\beta^\prime$. Using both (\ref{eq6}) and (\ref{eq7}), the distribution for (\ref{eq3}) is now split around $z=0$ as follows:
\begin{align}
\text{Pr}\left( F_\beta \right) &= \text{Pr}\left( F_\beta \leq z \right) = 1_{z \gt 0} \times \left[ \Phi(rz; r^\prime, \sigma^2) + \exp \left( \lambda r + \dfrac{\left( \frac{\lambda \sigma^2}{z} \right)^2 - \frac{2 r^\prime \lambda \sigma^2}{z}}{2 \sigma^2} \right) \right. \nonumber \\
& \left. \times  \left( 1 - \Phi \left( rz; \left( r^\prime - \frac{\lambda \sigma^2}{z} \right), \sigma^2  \right)   \right) \right] + 1_{z = 0} \times \Phi(0; r^\prime, \sigma^2)  \nonumber \\
&+ 1_{z \lt 0} \times \left[ \Phi(rz; r^\prime, \sigma^2) - \exp \left( \lambda r + \dfrac{\left( \frac{\lambda \sigma^2}{z} \right)^2 - \frac{2 r^\prime \lambda \sigma^2}{z}}{2 \sigma^2} \right) \right. \nonumber \\
& \left. \times \Phi \left( rz; \left( r^\prime - \frac{\lambda \sigma^2}{z} \right), \sigma^2  \right)    \right], \label{eq8}
\end{align}
where $\Phi(x; \mu, \sigma^2)$ denotes the standard normal or gaussian distribution at the value $x$ for a mean, $\mu$ and variance, $\sigma^2$. (Refer to Appendix \ref{secA3} for the proof). Similar to before the focus is on the indicator $1_{z \geq 0}$ as defined in (\ref{eq8}).
\begin{figure}
\centering
\begin{subfigure}[b]{0.45\textwidth}
         \centering
         \includegraphics[scale=0.15]{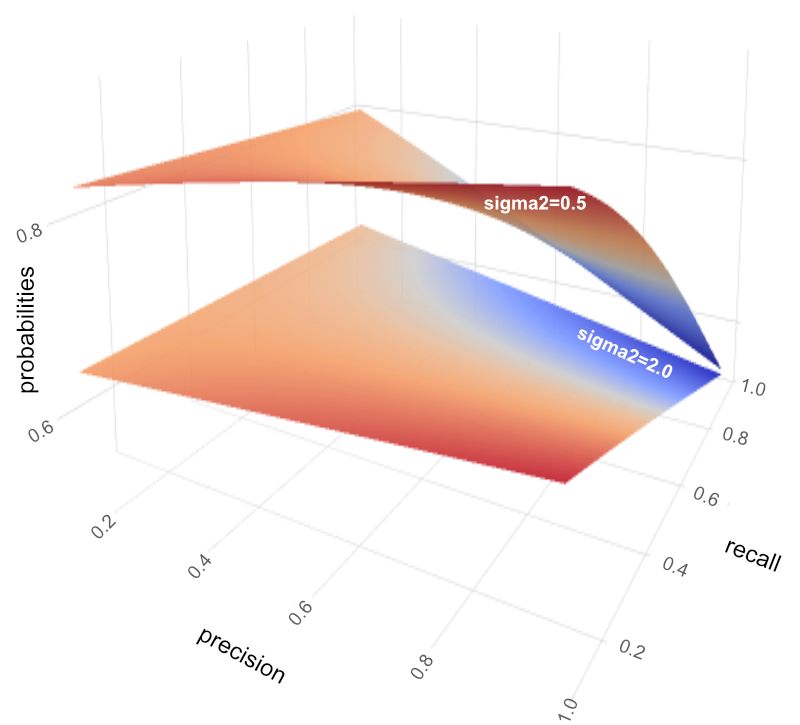}
         \caption{$P(F_\beta < 0.4)$ and $\lambda=0.5$}
         \label{fig2a}
\end{subfigure}
\begin{subfigure}[b]{0.45\textwidth}
         \centering
         \includegraphics[scale=0.15]{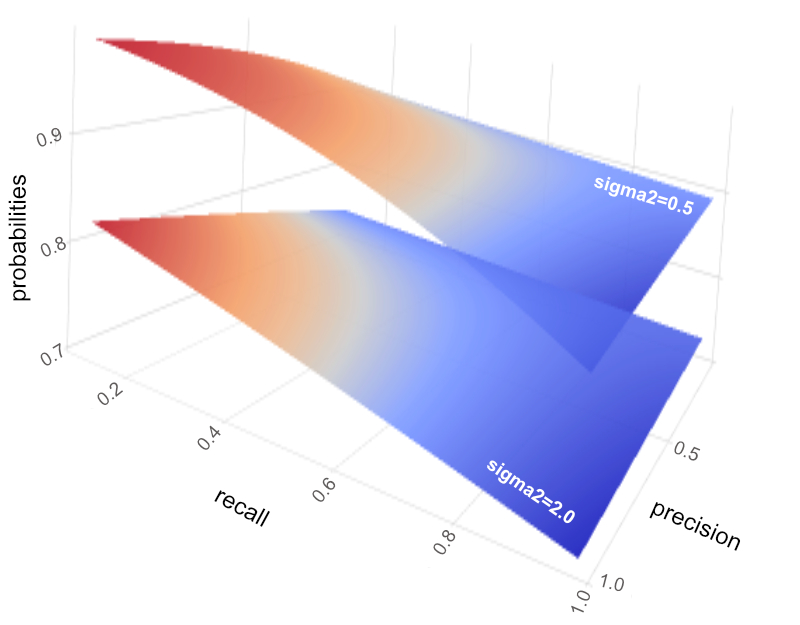}
         \caption{$P(F_\beta < 0.8)$ and $\lambda=0.5$}
        \label{fig2b}
\end{subfigure}
\begin{subfigure}[b]{0.45\textwidth}
         \centering
         \includegraphics[scale=0.15]{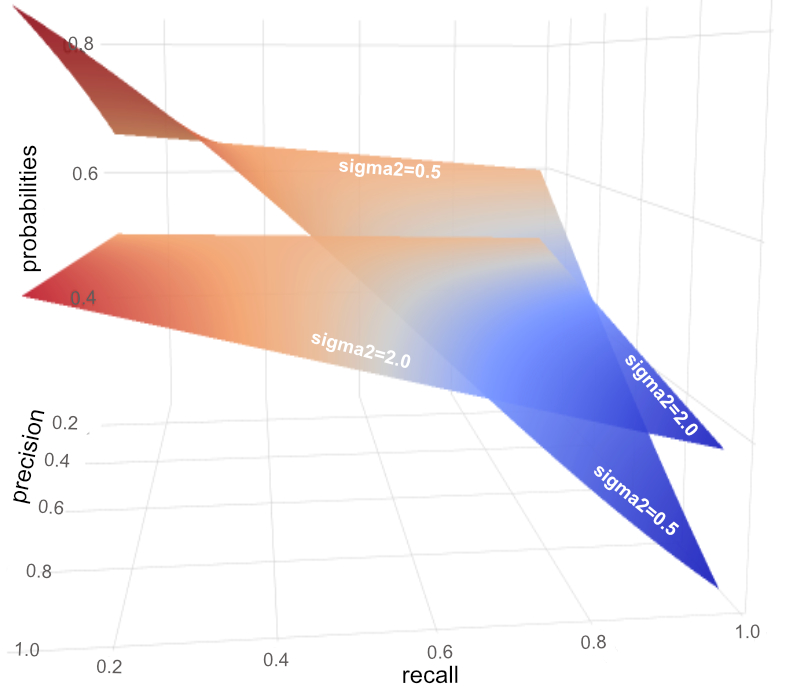}
         \caption{$P(F_\beta < 0.4)$ and $\lambda=2.0$}
         \label{fig2c}
\end{subfigure}
\begin{subfigure}[b]{0.45\textwidth}
         \centering
         \includegraphics[scale=0.15]{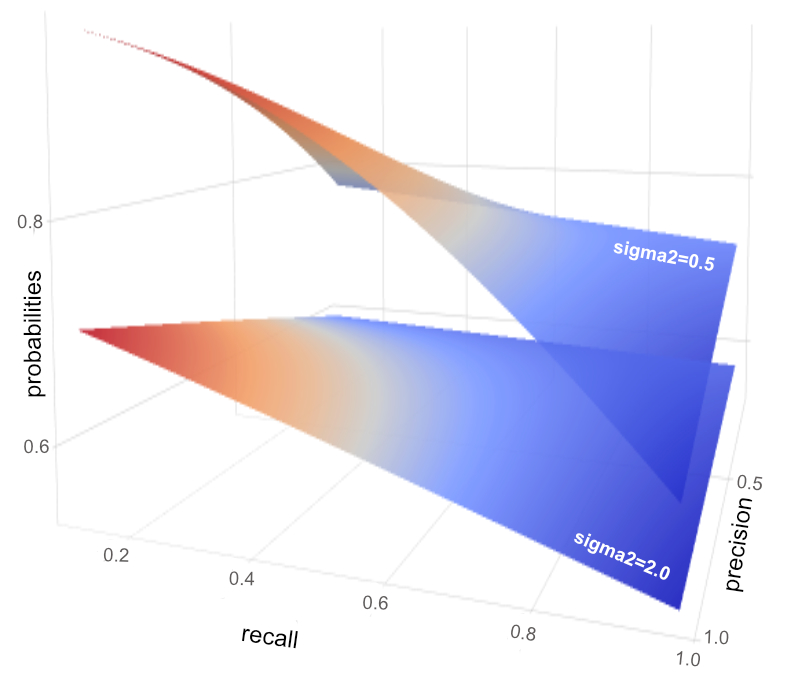}
         \caption{$P(F_\beta < 0.8)$ and $\lambda=2.0$}
         \label{fig2d}
\end{subfigure}
\caption{Probability Mass Surface: Ga \& IE for precision versus recall, $\beta=16$, $\lambda \in [0.5, 2.0]$, and $\sigma^2 \in [0.5, 2.0]$. The cumulative probability is computed for low ($0.4$) and high ($0.8$) values.} \label{fig2}
\end{figure}
Since this distributional mixture has more flexibility due to more parameters, Figure \ref{fig2} highlights this when $\lambda \in [0.5, 2.0]$, and $\sigma^2 \in [0.5, 2.0]$. The probabilities are computed again at a lower z value, $0.4$, and at a higher z value, $0.8$ for comparison. For a fixed $\lambda$, varying $\sigma^2$ impacts the curvature of the surface with higher $\sigma^2$ values producing a flattening effect. Figure \ref{fig2a} shows this distinctly. Conversely, as $\lambda$ increases with a fixed $\sigma^2$, the rate at which the surface changes is very apparent. This can be seen by juxtaposing Figure \ref{fig2c} and \ref{fig2a} or Figure \ref{fig2d} and \ref{fig2b} and noticing that the increase of $\lambda$ produces a clear increase in the rate of change. These observations match the intuition that $\sigma^2$ is linked to the shape of the bell curve, and $\lambda$ is linked to a rate of change. It also serves as a basis of intuition behind the algorithm in Section \ref{sec5}. That is, a faster rate of change along with a curved (and/or smoother) surface would provide loss penalties that adapt quickly per batch using the aggregated information from precision and recall.

\section{Knee algorithm and Weighted Cross Entropy}\label{sec5}
\subsection{Knee algorithm to find optimal $\beta$ values}\label{sub1sec5}
Now that probabilities, or $\text{Pr}(F_\beta \leq z)$ for some $z \in [0,1]$ are established in section \ref{sub1sec4} and \ref{sub2sec4}, the goal is to use them to get an optimal $\beta$ value, $\beta_{opt}$. There are a couple things to consider. First, because $\beta$ is grouped into $\beta^\prime$ and $\beta^{\prime\prime}$ with distributional assumptions, using maximum likelihood estimation (MLE) is not particularly suitable here. Also, $\beta_{max}$, $\sigma^2$, and $\lambda$ from (\ref{eq4}), (\ref{eq5}), and (\ref{eq7}) are set in advance and do not need to be estimated. Second, our observed data is only one data point per training batch, namely precision and recall. Given this and the natural bend of the $F_\beta$ function, a \textit{knee} algorithm is applicable. From \cite{bib4}, the knee of a curve is associated with good operator points in a system right before the performance levels off. This removes the need for complex system-specific analysis. Furthermore, they have provided a definition of curvature that supports their method being application independent -- an important property for this paper. Algorithm \ref{algo1} implements (and slightly alters) Kneedles algorithm from \citeauthor{bib4} to detect the knee in the $F_\beta$ curve. Refer to Algorithm \ref{algo1} for the formal pseudo code.
\begin{algorithm}
\caption{Calculate $\beta_{opt}$}\label{algo1}
\begin{algorithmic}[1]
\Require $n \gt 0 \wedge \beta_{max} \gt 0$
\Ensure $\beta_{opt} \gt 0$ 
\State Compute $p$, and $r$ from the training batch.
\State Initialize $b_{sn}, b_{d}, b_{lmx}, p_{sn}, p_{d}, p_{lmx}, p_s$ to empty arrays.
\State $b_s \Leftarrow [ b_{s_1}, ..., b_{s_n} ]$ where $b_{s_n} = \beta_{max}$
\For {($i=0$ to $n$)}
	\State $z \Leftarrow F_{\beta=b_{s_i}}^{n=1} (p, r)$ using eqn (\ref{eq2})
	\State $p_{s_i} \Leftarrow$ $\text{Pr}(F_\beta \leq z \vert \beta = b_{s_i}, p=p, r=r)$ using section \ref{sub1sec4} or  \ref{sub2sec4}
\EndFor
\If{$r \lt p$}
	\State $p_{max} = \max(p_s)$
	\For {($i=0$ to $n$)}
		\State $p_{s_i} \Leftarrow$ $p_{max} - p_{s_i}$
	\EndFor
\EndIf
\State$b_{max} = \max(b_s)$, $p_{max} = \max(p_s)$, $b_{min} = \min(b_s)$, $p_{min} = \min(p_s)$
\For {($i=0$ to $n$)}
	\State $b_{sn_i} \Leftarrow \dfrac{b_{s_i} - b_{min}}{b_{max}-b_{min}}$
	\State $p_{sn_i} \Leftarrow \dfrac{p_{s_i} - p_{min}}{p_{max}-p_{min}}$
	\State $b_{d_i} \Leftarrow b_{sn_i}$
	\State $p_{d_i} \Leftarrow p_{sn_i} - b_{sn_i}$
	\If {$(i \geq 1) \wedge (i \lt n)$}
		\If {$(p_{d_{i-1}} \lt p_{d_i}) \wedge (p_{d_{i+1}} \lt p_{d_i})$}
			\State $p_{lmx_i} \Leftarrow p_{d_i}$
			\State $b_{lmx_i} \Leftarrow b_{d_i}$
		\EndIf
	\EndIf
\EndFor
\If {$p_{lmx}$ is a non-empty array}
	\State $\beta_{opt} = \text{ mean}(p_{lmx})$
\Else
	\State $\beta_{opt} =1$ as per section \ref{sec3}
\EndIf
\end{algorithmic}
\end{algorithm}
A brief explanation in plain words is as follows:
\begin{enumerate}
\item For any training batch, compute precision (p) and recall (r). Then with a predefined $\beta_{max}$ value, set $n$ equally spaced values, $b_{s_i}$, up to $\beta_{max}$, and use section \ref{sec4} to compute $p_{s_i} = \text{Pr}(F_\beta \leq z \vert \beta = b_{s_i}, p=p, r=r)$. (This replaces step 1 from \citeauthor{bib4}). Let $D_s$ represent this smooth curve as $D_s = \{(b_{s_i}, p_{s_i}) \in \mathbb{R}^2 \vert b_{s_i}, p_{s_i} \geq 0\}$ for $i = 1,...n$.
\item When $r \lt p$, convert to a knee by taking the difference of the probabilities from the maximum. That is, $p_{s_i} = \max(p_{s}) - p_{s_i}$ for $i = 1,...n$. This is necessary because of the formulation of the $F_\beta$ metric. 
\item Normalize the points to a unit square and call these $b_{sn}$ and $p_{sn}$.
\item Take the difference of points and label that $b_d$ and $p_d$.
\item Find the candidate knee points by getting all local maxima's, label that $b_{lmx}$ and $p_{lmx}$.
\item Take the average of $p_{lmx}$ and this will be $\beta_{opt}$. (This simplifies \citeauthor{bib4}).
\end{enumerate}

\subsection{Proposal Weighted Binary Cross-Entropy}\label{sub2sec5}
The weighted binary cross-entropy loss is primarily focused around the imbalanced use case where a minority class exists. This paper posits that from the shuffling of data observations, as is frequently done while training, relevant aggregate information is available to use from the batch. For instance, say for a fixed minority class observation, $y$, it is grouped among different batches of the majority class. The interaction effect of $y$ among these randomly varying training batches is often overlooked. It is this interaction that can be inferred through the precision and recall aggregates, then transferred as a penalty to the loss function via $\beta_{opt}$ in a probabilistic way. By using Algorithm \ref{algo1} to get $\beta_{opt}$, the proposed loss is, 
\begin{align}
L(\textbf{f(x; $\theta$)} \vert \beta^2_{opt}, \textbf{x}) &= -\sum_i \left \{  y_i \log \left(  f_i(\textbf{x}; \theta) \right) + \left( 1-y_i \right) \times \log \left( 1 - f_i(\textbf{x}; \theta) \right) \right. \nonumber \\
& \left. \times \left( \frac{1_{\{[1 - f_i(\textbf{x}; \theta)] \leq 0.5}\}}{1+\beta^2_{opt}} + (1+\beta^2_{opt}) \times 1_{\{[1 - f_i(\textbf{x}; \theta)] \gt 0.5\}} \right) \right \}, \label{eq9}
\end{align}
where the function $f_i(\textbf{x}; \theta)$ is the $i$-th element from the prediction of a neural network using the inputs \textbf{x} and training weights, $\theta$; and $y_i$ is the $i$-th element of the true target label. When considering the majority class, or $y_i=0$ for $i=\{1,...,m\}$, the loss is weighted by $(1+\beta^2_{opt})$. Therefore, for correctly predicted observations, the loss has a reduction by $(1+\beta^2_{opt})$. When incorrectly predicted, the loss is magnified by the same amount. For the minority class, or $y_i=1$ for $i=\{1,...,n-m\}$, the loss is unchanged. This is intentional because under imbalanced data there are far less observations, and computing precision and recall lead to numerical instability or frequent edge cases for Algorithm \ref{algo1}. 

\subsection{Understanding the $\beta_{opt}$ Surface and Weighted Cross-Entropy}\label{sub3sec5} 
\begin{figure}
\centering
\begin{subfigure}[b]{0.45\textwidth}
         \centering
         \includegraphics[scale=0.15]{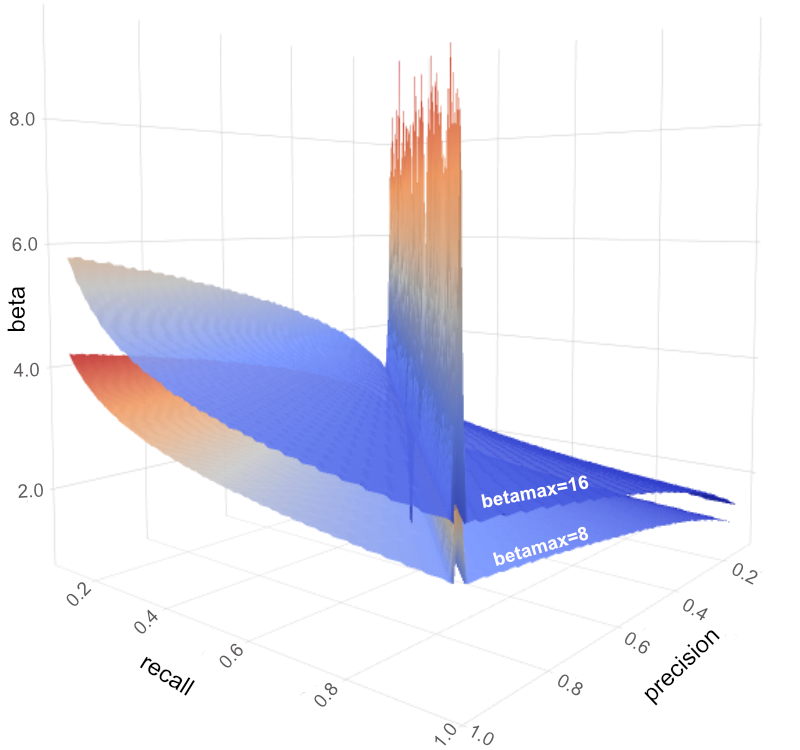}
         \caption{U \& IU for $\beta_{max} \in [8, 16]$}
         \label{fig3a}
\end{subfigure}
\begin{subfigure}[b]{0.45\textwidth}
         \centering
         \includegraphics[scale=0.15]{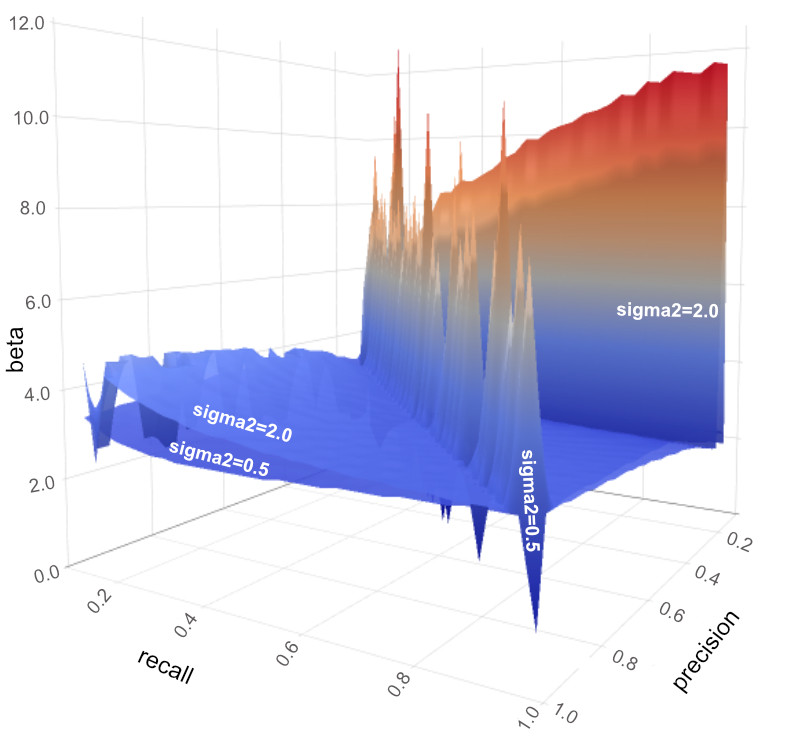}
         \caption{Ga \& IE for $\beta_{max}=16$,  $\lambda=0.5$}
        \label{fig3b}
\end{subfigure}
\begin{subfigure}[b]{0.45\textwidth}
         \centering
         \includegraphics[scale=0.15]{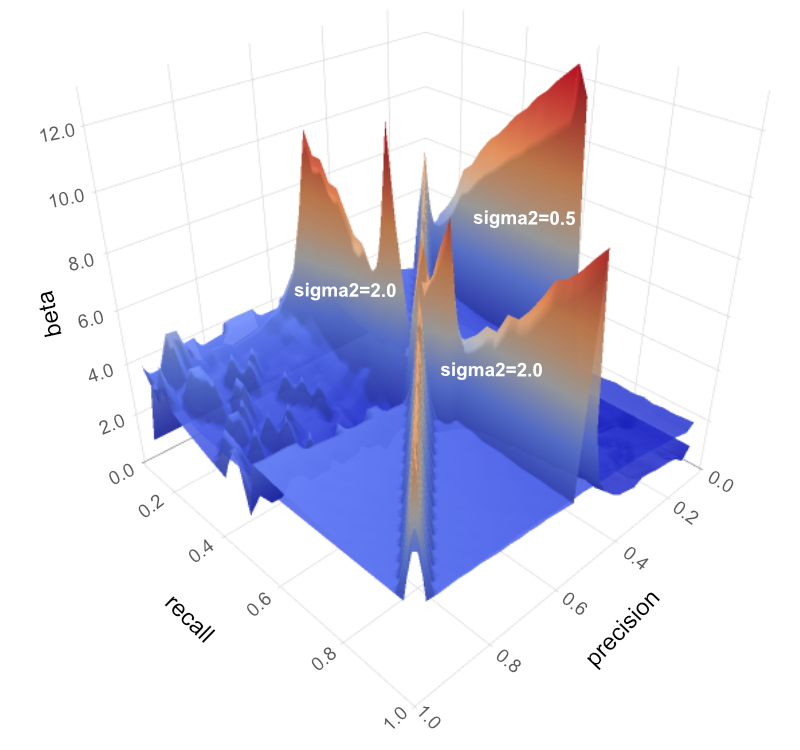}
         \caption{Ga \& IE for $\beta_{max}=16$, $\lambda=2.0$}
        \label{fig3c}
\end{subfigure}
\caption{$\beta_{opt}$ Surface: (a) U \& IU where $\beta_{max} \in [8, 16]$, (b) and (c) Ga \& IE for fixed $\beta_{max}=16$, $\lambda \in [0.5,2.0]$ respectively. Note for Algorithm \ref{algo1} $n=300$ equally spaced points.} \label{fig3}
\end{figure}
Figure \ref{fig3} highlights the surface generated from Algorithm \ref{algo1} leveraging probabilities from Section \ref{sec4}. First, the U \& IU mixture or Figure \ref{fig3a} suggests that the shape of the surface remains relatively similar even when doubling $\beta_{max}$. This is an important point toward fixing $\beta_{max}=16$ for the Ga \& IE. Based on Figure \ref{fig3a}, the U \& IU mixture penalizes more on the outskirts of recall, while the immediate penalties arise on the diagonal of the unit square. This suggests that precision and recall estimates from training cause immediate penalties when they are on opposite ends of the $[0,1]$ range as well as on the diagonal when these values start to even out. For Ga \& IE mixture, Figures \ref{fig3b} and \ref{fig3c} show some similar conclusions as U \& IU mixture along with additional insights. For a lower rate of $\lambda=0.5$, or Figure \ref{fig3b}, diagonal spikes for $\sigma^2 \in [0.5, 2.0]$ as well as a precision centric penalty for higher $\sigma^2$ (i.e., $\sigma^2=2$) are seen. For a higher rate of $\lambda=2.0$, or Figure \ref{fig3c}, a similar diagonal is retained for $\sigma^2 \in [0.5, 2.0]$ as in Figure \ref{fig3b}. Furthermore, for increasing values of $\sigma^2$, the penalty evolves from precision centric to a vertical separation on the unit grid at around a precision of $0.4$. The overall interpretation is the following: for a lower $\lambda$, increasing $\sigma^2$ creates a slightly more precision based penalty; while for a higher $\lambda$, increasing $\sigma^2$ causes the penalty to become more balanced between recall and precision. The choice of these parameters are problem specific, but provides the practitioner flexibility in determining the best selection for their use case. On a separate note, from Figure \ref{fig3}, a spiky surface is obvious, which is partially explained by having a default setting in the algorithm. This is a strong sign of immediate and configurable penalties.

\section{Datasets and Experimentation}\label{sec6}
\subsection{Datasets}\label{sub1sec6}
The origins of the $F_\beta$ metric come from text retrieval, so it is important to verify this method across different categories of data. In particular, image data from CIFAR-10\footnotemark[1], text data from IMDB movie sentiments \cite{bib5} and structured/tabular data from the Census Income Dataset \cite{bib6} are tested. For each experiment, the primary label (i.e., label 1) is either imbalanced or forced to be imbalanced to reflect real world scenarios. Because CIFAR-10 contains multiple image labels, the airplane label is the primary label and all others are combined. This yields a 10\% class imbalance. IMDB movie sentiment reviews (positive/negative text) are not imbalanced. The positive sentiments in the training data are reduced to 1K randomly sampled sentiments yielding a 7.4\% imbalance (Table \ref{tab2}). The Census Income Tabular data contains 14 input features (i.e., age, work class, education, occupation, etc) with 5 numerical and 9 categorical features. The binary labels are greater than 50K salary (label 1) and less than 50K salary (label 0). By default, greater than 50K salary is already imbalanced at 6.2\%. The training and validation dataset sizes for each data category are as follows: for CIFAR-10, 50K training and 10K validation, for IMDB, 13.5K training and 25K validation, and for the Census data, 200K training and 100K validation. In terms of class imbalance, this paper considers an imbalance or proportion of label 1 under 10\% to be significantly imbalanced and between 10\% to 25\% to be moderately imbalanced. Some heuristic rationale for a 10\% imbalance, a model that has perfect recall or 100\%, a precision of $p=\frac{1}{3}$ would be required to get $F_1=0.5$. In practical examples, this scenario can occur with weakly discriminative features. Therefore, this paper seeks to test this algorithm in scenarios that would need an improved precision. A variety of imbalanced and balanced scenarios will be tested in this paper. \\

Two real-life use cases related to cylindrical tanks are also considered, providing a physical domain to test Algorithm \ref{algo1}. \cite{bib103} developed an arithmetic optimizer with slime mould algorithm and \cite{bib104} developed an evolutionary based algorithm with slime mould algorithm; both algorithms focus on global parameter optimization. They tested these algorithms on several benchmark problems, one of which is called the pressure vessel design. The problem is a constrained parameter optimization (i.e., material thickness, and cylinder dimensions) for minimizing a cost function. This paper focuses on using the HAOASMA algorithm by \citeauthor{bib103} in a simulation to convert the problem into a binary classification. The second use case is derived from Underground Storage Tanks (UST) and is also inspired by \citeauthor{bib103}'s pressure vessel problem. The physical shape (i.e., the cylindrical shape) of USTs is similar to the pressure vessel design. USTs are used to store petroleum products, chemicals, and other hazardous materials underground. These structures could deform underground and possibly explain a false positive leak. \cite{ramdh1} and \cite{ramdh2} explored parameter optimization of UST dimensions changing from cylindrical to ellipsoidal. The observed data are vertical (underground) height measurements, which can contain uniformly distributed error. \citeauthor{ramdh2} used these measurements and the volumetric equations (\ref{fc}), (\ref{fch}), (\ref{fed}), and (\ref{fedh}) - derived from a cross-sectional view - to develop a methodology to estimate tank dimensions and test if the shape has deformed. The cross-sectional view can be seen in Figure \ref{histogram}. \\

The conversion of both of these real-life use cases into a classification involves establishing a baseline set of parameters to simulate data for label 0. Varying these parameters will allow simulation of data for label 1. For the pressure vessel design, the baseline parameters from HAOASMA are $T_s=1.8048$, $T_h=0.0939$, $R=13.8360$, and $L=123.2019$. To convert this to a classification, the parameters for thickness $T_s$ and $T_h$ are changed from the baseline while $R$ and $L$ dimensions are drawn from a normal distribution. By using values for $T_s$, $T_h$, $R$ and $L$, the cost function is computed using \ref{pveq}. These cost values concatenated with $R$ and $L$ arrays serve as the input to a neural network classifier. The label 1 reflect simulated data using $T_s$ and $T_h$ that are changed from the baseline. These variations are  $T_s=1.7887 $ and $T_h \in \{0.0313, 0.2817\}$. Label 0 is HAOASMA baseline values $T_s = 1.8048$ and $T_h = 0.0939$. Appendix \ref{secA4} provides the equations, the distributional plots seen in Figure \ref{pv}, and a detailed explanation of the simulation procedure (Algorithm \ref{algo2}). For the UST problem, \citeauthor{ramdh1} used a measurement error model with an error on the height measurement and another on the volume computation. The same model is used to simulate data in this paper. The baseline is a cylinder and the variations to vertical and horizontal axes $a$ and $b$ represent a deformed cylinder to an ellipse. By using $r$, $L$, $h$, $a$ and $b$ along with (\ref{fc}) and (\ref{fed}) or (\ref{fch}) and (\ref{fedh}) the volume is computed. These volumes concatenated with noisy height measurements are the inputs to a neural network classifier. The label 1 reflect simulated data using the variations to the cylinder or the baseline. These variations are $a \in \{3.2, 3.8\}$ and $b \in \{5.0, 4.2105\}$. Label 0 will be the baseline cylinder with radius $r=4$ and the length of $L=32$. Refer to Appendix  \ref{secA5} for detailed explanation of the simulation Algorithm \ref{algo3} along with comparison plots and volume equations.

\footnotetext[1]{https://www.cs.toronto.edu/~kriz/cifar.html}

\subsection{Model Networks}\label{sub2sec6}
\subsubsection{Image Network}\label{sub2sub1sec6}
For the CIFAR-10 image dataset, ResNet (\citealt{bib7}) version 1 is applied. The number of layers for ResNet is 20, which upon initial experimentation is adequate for speed and generalization in this case. Adam optimizer was implemented with a learning rate of $1e^{-3}$ with total epochs of 30. No learning rate schedule because of the intentional lower number of epoch in order to validate faster training via this proposed loss algorithm. The training batch size is 32. Modest data augmentation is done -- random horizontal and vertical shifts of 10\%, and horizontal and vertical flips. 

\subsubsection{Text Network}\label{sub2sub2sec6}
For the IMDB movie sentiments, a Transformer block (which applies self-attention \citealt{bib9}) is used. The token embedding size is 32, and the transformer has 2 attention heads and a hidden layer size of 32 including dropout rates of 10\%. A pooling layer and the two layers that follow -- a dense RELU activated layer of size 20 and a dense sigmoid layer of size 1 -- give the final output probability. As for preprocessing, a vocabulary size of 20K and maximum sequence length of 200 is implemented. The training batch size is 32.

\subsubsection{Structured/Tabular Network}\label{sub2sub3sec6}
For the Census Income Dataset, a standard encoder embedding paradigm \cite{bib10} is used. Specifically, all categorical features with an embedding size of 64 are concatenated, then numerical features are concatenated to this embedding vector. Afterwards, a 25\% dropout layer and the two layers that follow -- a fully connected dense layer with GELU activation of size 64 and a sigmoid activated layer of size 1 -- provide the final output probability. The training batch size is 256.

\subsubsection{UST/Vessel Network}\label{sub2sub4sec6}
For the real life use cases on simulated data the model network is simple because of the minimal amount of features. The network is a sequential set of dense layers of sizes 20, 10 and 1. The last layer of size 1 has a sigmoid activation to give the final output probability. Additionally, a drop out of 10\% is added after both middle layers. The training batch size is 128.

\subsection{Experimental Results}\label{sub2sec6}
The results in Table \ref{tab2} compare the use of the loss function (\ref{eq9}) by different models based on U \& IU and Ga \& IE to a baseline case of ordinary cross-entropy. All results shown in this table are computed on the validation datasets for each data category above (see above for the dataset sizes). For ease of presentation, $M_1^\beta$ is Model 1: U \& IU from (\ref{sub1sec4}). $M_2^{(\lambda, \sigma^2)}$ is Model 2: Ga \& IE from (\ref{sub2sec4}). The superscripts $\beta$ and $(\lambda, \sigma^2)$ are the parameters being explored. $M_B$ is the baseline or the same model network that is trained using ordinary cross-entropy.

\subsubsection{Image Results}\label{sub2sub1sec6}
For the image network, Table \ref{tab2} shows modest improvement over the baseline under the $M_1^\beta$ for a moderately sized $\beta=8$. This suggests that image data trains better under constant penalties on the outskirts of the unit square toward the imbalance of high precision and low recall. High precision and low recall imply image confusion between classes in the feature embedding space. In fact, this can lead to large implications as in \cite{bib13}. Algorithms like DeepInspect \cite{bib11} help to detect confusion and bias errors to isolate misclassified images leading to \textit{repair based} training algorithms such as \cite{bib15} and \cite{bib14}. But \cite{bib12} empirically shows that such repair or de-biasing algorithms can be inaccurate with one fixed-seed training run. The importance of the $M_1^\beta$ result is now evident because $M_1^\beta$ quickly penalizes the network in a way that inherently mirrors algorithms like DeepInspect's confusion/bias detection without the need for repair algorithms. 

\subsubsection{Text Results}\label{sub2sub2sec6}
The training results for the text network by far show the most improvement with a nearly 14\% boost in the $F_1$ score over the baseline for the $M_2^{(\lambda, \sigma^2)}$ model. Not only is the performance notable, the model parameter selections are consistent -- the parameters move in the same direction. In other words, given the parameters $\lambda=0.5$ and $\sigma^2=0.5$, the training shows improvement over the baseline and this improvement continues in the same direction when $\lambda=0.01$ and $\sigma^2=0.01$. This is similar to section \ref{sub2sub1sec6} because first, the architecture is generalizing better (seen by the $F_1$ score) for label confusion (i.e., language context) and second, it adjusts for intentionally configured imbalance and incorrect labeling (a known issue for this dataset). The incorrect labeling in the IMDB dataset is shown to be non-negligible -- upwards of 2-3\% -- by \cite{bib17} and \cite{bib16}. In particular, \citeauthor{bib16} show that small increases in label errors often cause a destabilizing effect on machine learning models for which \textit{confident learning} methodology is developed to detect them. \citeauthor{bib17} analyze 18 methods (including \textit{confident learning}) for Automated Error Detections (AED) and shows the importance of AED for data cleaning. In close proximity to the AED methodology, another paradigm is Robust Training with Label Noise. \cite{bib18} provides an exhaustive survey ranging from robust architectures (i.e., noise adaptation layers) and robust regularization (i.e., explicit and implicit regularization) to robust loss (i.e., loss correction, re-weighting, etc.) and sample selection. It is in this context that the $M_2^{(\lambda, \sigma^2)}$ framework sits between AED and Robust Training with Label Noise on this IMDB dataset which is known to have errors. $M_2^{(\lambda, \sigma^2)}$ serves two purposes: (1) as a robust loss through the $\beta_{opt}$ re-weighting on the batch and, (2) as a means to detect and down weight possible label errors. 

\subsubsection{Structured/Tabular Results}\label{sub2sub2sec6}
The results for the structured/tabular network do not show any $F_1$ improvement over the baseline nor any indication of possible improvement through the extra parameter variations. From Table \ref{tab2}, the best performing model for this dataset (not the baseline) is $M_2^{(\lambda, \sigma^2)}$ where $\lambda=2.0$ and $\sigma^2=0.5$. The interpretation of this parameter configuration suggests that training tabular data is very susceptible to both low precision and recall, hence the high penalty in that area of the unit square in Figure \ref{fig3}. Despite embedding categories and numeric features into a richer vector space, the non-contextual nature of tabular data may not necessarily be best trained through these architectures. Furthermore, \cite{bib19} applies a two dimensional embedding (i.e., simulating an image) to this Census dataset and the results show that a decision tree (i.e., xgboost) would perform similarly. It is worth mentioning that \citeauthor{bib19} present these results with an accuracy measure (not $F_1$) which is misleading since the data is naturally imbalanced. However, a similar general conclusion is given by \cite{bib20} for tabular data -- decision trees have faster training time and generally comparable accuracy as compared with embedding based architectures. These results are unsurprising because as stated by \cite{bib21} tabular data is not contextually driven data like images or languages which contain position-related correlations. It is heartening to notice, that after \citeauthor{bib21} apply a casually aware GAN to the census data, the resulting $F_1$ score ($0.509$) is similar to the baseline result in Table \ref{tab2} ($0.5193$). Because of these results, there is an important finding: the type of data, in particular contextual data which is the basis for the creation of the $F_\beta$ metric, plays a significant role when using the metric alongside a loss function. This hypothesis is studied further in the benchmark data in Section \ref{sub3sec1}.

\subsubsection{UST/Pressure Vessel Results}\label{sub2sub4sec6}
The results for the simulation of real life use cases can be found in Table \ref{tab_bench_real}. In the UST case, it is evident that this methodology outperforms the baseline cross-entropy in determining a shape change from a cylinder to an ellipse. For example, in the easier scenario for CvE ($a=3.2$) the $M_2^{(\lambda, \sigma^2)}$ model family appears to be better. However, in cases of the extra variations, the $M_1^8$ and  $M_2^{(0.01,0.01)}$ perform the same. This trend is also observed in the results for the Image and Text data presented in Table \ref{tab2}. The interpretation is that a slightly more recall-centric penalty may be optimal for this scenario. Interestingly, for the easier CHvEH scenario ($a=3.2$), the $M_2^{(\lambda, \sigma^2)}$ model family also appears to be better, and the extra variations for  $M_1^{32}$ and  $M_2^{(5.0,5.0)}$ perform the same. These variations mirror CvE but in the other direction, suggesting that a balanced or slightly more precision-centric penalty is optimal. In the difficult scenario ($a=3.8$), both CvE and CHvEH are closely aligned with $M_2^{(\lambda, \sigma^2)}$ model family. For CHvEH the best performer is the $M_1^{32}$ variation. Overall, there is between 12\% to 28\% improvement over the baseline or standard cross-entropy for this simulation. Regarding the PV data, for the easier scenario ($t_h=0.0313$) the $M_1^\beta$ family appears to be better, with the $M_2^{(\lambda, \sigma^2)}$ model family not far behind. In the difficult scenario ($t_h=0.2817$) there is no improvement over the baseline cross-entropy but the best performing model family is $M_2^{(\lambda, \sigma^2)}$. The reason is likely due to the significant overlap in distribution seen in Figure \ref{pv}. These results are impactful because the commonality between model families begins to surface. For the easier scenario, a more recall-centric penalty turns out to be better, while in the difficult scenario, a balanced or slightly precision-centric penalty is more effective. This finding is intuitive.This finding is intuitive.

\begin{sidewaystable}
\sidewaystablefn%
\begin{center}
\begin{minipage}{\textwidth}
\caption{Best F1 Score for $M_1^\beta$ and $M_2^{(\lambda, \sigma^2)}$ over $30$ Epochs}\label{tab2}
\begin{tabular*}{\textwidth}{@{\extracolsep{\fill}}lcccccccccc@{\extracolsep{\fill}} }
\toprule
&\multicolumn{1}{@{}c@{}}{Baseline} & \multicolumn{5}{@{}c@{}}{Parameter Variations Section \ref{sub3sec5}} & \multicolumn{4}{@{}c@{}}{Extra Variations}  \\ \cmidrule{2-2}\cmidrule{3-7}\cmidrule{8-11}
Dataset     & $M_B$     & $M_1^{16}$ & $M_2^{(0.5, 0.5)}$ & $M_2^{(0.5, 2.0)}$ & $M_2^{(2.0, 0.5)}$ & $M_2^{(2.0, 2.0)}$     & $M_1^{8}$ & $M_1^{32}$ &  $M_2^{(0.01, 0.01)}$ &  $M_2^{(5.0, 5.0)}$ \\
\midrule
Image\footnotemark[1]          & 0.8161                 & \textbf{0.8261} & 0.8085 & 0.8193 & 0.8232 & 0.8257       & \textbf{0.8266} & 0.8068 & 0.8087 & 0.8178 \\
Text\footnotemark[2]             & 0.6749                 & 0.6393 & \textbf{0.7175} & 0.6170 & 0.6547 & 0.6673       & 0.7236 & 0.5460 & \textbf{0.7666} & 0.7364 \\
Structured\footnotemark[3]   & \textbf{0.5193}     & 0.4170 & 0.3917 & 0.4126 & 0.4635 & 0.3930                   & 0.3824 & 0.3511 & 0.3890 & 0.4516 \\
\botrule
\end{tabular*}
\footnotetext[1]{The image dataset is the CIFAR10. The airplane label versus the remaining labels is the binary label basis. It gives a training data imbalance of 10\%. Training data size is 50K and validation is 10K.}
\footnotetext[2]{The text dataset for NLP is the IMDB movie sentiment with binary label of positive/negative sentiment. The vocabulary size is 20K and the maximum review length is 200. The training set is imbalanced by choosing only 1K positive sentiments which yields an imbalance of 7.4\%. The training data size is 13.5K and validation is 25K.}
\footnotetext[3]{The structured or tabular data set is the Census Income Dataset from UCI repository. The labels are greater than or less than 50K salary. The data is already imbalanced with a rate of 6.2\% for $>$50K. The training data size is 200K and the validation is 100K.}
\end{minipage}
\end{center}
\end{sidewaystable}

\begin{sidewaystable}
\sidewaystablefn%
\begin{center}
\begin{minipage}{\textwidth}
\caption{Best F1 Score for $M_1^\beta$ and $M_2^{(\lambda, \sigma^2)}$ over $30$ Epochs}\label{tab_bench_real}
\begin{tabular*}{\textwidth}{@{\extracolsep{\fill}}lcccccccccc@{\extracolsep{\fill}} }
\toprule
&\multicolumn{1}{@{}c@{}}{Baseline} & \multicolumn{5}{@{}c@{}}{Parameter Variations Section \ref{sub3sec5}} & \multicolumn{4}{@{}c@{}}{Extra Variations}  \\ \cmidrule{2-2}\cmidrule{3-7}\cmidrule{8-11}
Dataset\footnotemark[0]  & $M_B$     & $M_1^{16}$ & $M_2^{(0.5, 0.5)}$ & $M_2^{(0.5, 2.0)}$ & $M_2^{(2.0, 0.5)}$ & $M_2^{(2.0, 2.0)}$     & $M_1^{8}$ & $M_1^{32}$ &  $M_2^{(0.01, 0.01)}$ &  $M_2^{(5.0, 5.0)}$ \\
\midrule
CvE\footnotemark[1]  &0.9691 &0.9228 &\textbf{0.9983} &0.9915 &0.9898 &0.9565 &\textbf{0.9966} &0.9915 & \textbf{0.9966} &0.9673 \\
CvE\footnotemark[2] &0.3169 &0.3147 &0.3351 &\textbf{0.3469} &0.3296 &0.3333 &0.3224 &0.3401 &0.3362 &\textbf{0.3573} \\
CHvEH\footnotemark[3] &0.9831 &0.9813 &0.9813 &0.9898 &\textbf{0.9915} &0.9831 &0.9726 &\textbf{0.9882} &0.9831 &\textbf{0.9882} \\
CHvEH\footnotemark[4] &0.2891 &0.3345 &0.3427 &\textbf{0.3515} &0.3262 &0.3159 &0.3636 &\textbf{0.3701} &0.3395 &0.3425 \\
PV\footnotemark[5] &0.9967 &\textbf{0.9992} &0.9983 &0.9483 &0.9831 &0.9967 &0.9967 &0.9891 &0.9727 &0.9958 \\
PV\footnotemark[6] &\textbf{0.7515} &0.4552 &0.5057 &0.4893 &0.4722 &0.5248 &0.4934 &0.4861 &0.4675 &0.5161 \\
\botrule
\end{tabular*}
\footnotetext[0] {The simulations for UST (Underground Storage Tanks) are for the cylinder versus ellipse (CvE) or cylinder with hemispherical end-caps versus ellipsoidal with hemi-ellipsoidal end-caps (CHvEH). For the PV or pressure vessel, the simulation is between varying thickness of the surface and head. Refer to Appendix \ref{secA4} and \ref{secA5} for details. The label 1 proportion is 25\% and total training and testing size is 1200. }
\footnotetext[1] {The simulation has label 0 with $r=4$ and $L=32$ versus label 1 of $a=3.2$ and $b=5.0$.}
\footnotetext[2] {The simulation has label 0 with $r=4$ and $L=32$ versus label 1 of $a=3.8$ and $b=4.2105$.}
\footnotetext[3] {The simulation has label 0 with $r=4$ and $L=32$ versus label 1 of $a=3.2$ and $b=5.0$.}
\footnotetext[4] {The simulation has label 0 with $r=4$ and $L=32$ versus label 1 of $a=3.8$ and $b=4.2105$.}
\footnotetext[5] {The simulation has label 0 with $t_s=1.8048$ and $t_h=0.0939$ versus label 1 of $t_s=1.7887$ and $t_h=0.0313$.}
\footnotetext[6] {The simulation has label 0 with $t_s=1.8048$ and $t_h=0.0939$ versus label 1 of $t_s=1.7887$ and $t_h=0.2817$.}
\end{minipage}
\end{center}
\end{sidewaystable}

\newpage

\subsection{Further Experimentation: Benchmark Analysis}\label{sub3sec1}
Following the benchmark analysis from \citeauthor{bib102}, a similar approach is done for the Image, Text, and Tabular data. This expands the analysis from Table \ref{tab2} to provide a more detailed and comprehensive view across various well-known datasets. The results can be found in Table \ref{tab_bench_cv}, \ref{tab_bench_nlp}, and \ref{tab_bench_tab}. The footnotes in these tables are explained as follows: the breakdown of train and test data sizes, the proportion of label 1, the labeling convention for label 1 versus label 0 (if multiple labels exist), and the location of the data, if necessary. For example, label 9 vs all means the label 9 is the label 1 and everything else is marked as label 0. Detailed explanations, links, and training details for all the datasets are provided in the footnotes for each table. At a high level, for images CIFAR-10, CIFAR-100 and Fashion MNIST are analyzed. For text, AG's News Corpus, Reuters Corpus Volume 1, Hate Speech and Stanford Sentiment Treebank are analyzed. For the tabular data, 10 classical datasets from UCI repository are analyzed. Finally, the same model networks from Section \ref{sub2sec6} will be used.

\subsubsection{Image Results}\label{sub3sub1sec1}
Comparing the CIFAR-10 result in Table \ref{tab2} versus \ref{tab_bench_cv}, the model family changes from $M_1^\beta$ to $M_2^{(\lambda, \sigma^2)}$. The interpretation remains the consistent: a recall centric based penalty is favored. The CIFAR-100 examples, with an imbalance of 1\%, follow a similar recall centric penalty for $M_1^{16}$ under the label convention 9 vs all. However, under the labeling 39 versus all, a more precision centric penalty is preferred. This illustrates the problem-specific nature of selecting a model family and parameter, showcasing the flexibility of this paper's methodology. Notably, there is a 14\% increase in the $F_1$ score for CIFAR-100 under the 39 versus all label convention. Fashion MNIST favors the $M_2^{(\lambda, \sigma^2)}$ with a more precision centric penalty. The most intriguing result is that, for all the extra variations, $M_2^{(5.0, 5.0)}$ is the most frequent performer, which is a more balanced penalty. This suggests that $M_2^{(5.0, 5.0)}$ could be a starting point of exploration given the balanced nature of the penalty distribution.

\begin{sidewaystable}
\sidewaystablefn%
\begin{center}
\begin{minipage}{\textwidth}
\caption{Best F1 Score for $M_1^\beta$ and $M_2^{(\lambda, \sigma^2)}$ over $30$ Epochs}\label{tab_bench_cv} 
\begin{tabular*}{\textwidth}{@{\extracolsep{\fill}}lcccccccccc@{\extracolsep{\fill}} }
\toprule
&\multicolumn{1}{@{}c@{}}{Baseline} & \multicolumn{5}{@{}c@{}}{Parameter Variations Section \ref{sub3sec5}} & \multicolumn{4}{@{}c@{}}{Extra Variations}  \\ \cmidrule{2-2}\cmidrule{3-7}\cmidrule{8-11}
Dataset\footnotemark[0]  & $M_B$     & $M_1^{16}$ & $M_2^{(0.5, 0.5)}$ & $M_2^{(0.5, 2.0)}$ & $M_2^{(2.0, 0.5)}$ & $M_2^{(2.0, 2.0)}$     & $M_1^{8}$ & $M_1^{32}$ &  $M_2^{(0.01, 0.01)}$ &  $M_2^{(5.0, 5.0)}$ \\
\midrule
CIFAR-10\footnotemark[1] &0.9216 &0.9088 &0.9204 &0.9196 &\textbf{0.9263} &0.9122 &0.9194 &0.9119 &\textbf{0.9268} &0.9173 \\
CIFAR-100\footnotemark[2] &0.7345 &\textbf{0.7804} &0.7273 &0.6941 &0.7594 &0.7692 &0.7501 &0.7167 &0.7314 &\textbf{0.7683} \\
CIFAR-100\footnotemark[2]  &0.6021 &0.6592 &0.6778 &\textbf{0.6871} &0.6381 &0.6818 &0.6509 &0.6351 &0.6702 &\textbf{0.6704} \\
Fashion MNIST\footnotemark[3]  &0.8651 &0.8638 &\textbf{0.8663} &0.8462 &0.8651 &0.8593 &0.8544 &0.8558 &0.8638 &\textbf{0.8672} \\
Fashion MNIST\footnotemark[3]  &0.9627 &0.9621 &0.9656 &0.9656 &\textbf{0.9675} &0.9648 &0.9615 &0.9648 &0.9641 &\textbf{0.9681} \\
\botrule
\end{tabular*}
\footnotetext[0] {All the datasets are easily found in Keras repository. The link is : \url{https://keras.io/api/datasets/}. }
\footnotetext[1]{Train/test 50K/10K, label 1 10\%, labeling is 1 vs all.  }
\footnotetext[2]{Train/test 50K/10K \& 50K/10K, label 1 1\% \& 1\%, labeling is 9 vs all \& 39 vs all.  }
\footnotetext[3]{Train/test 50K/10K \& 50K/10K, label 1 10\% \& 10\%, labeling is 0 vs all \& 9 vs all.  }
\end{minipage}
\end{center}
\end{sidewaystable}

\subsubsection{Text Results}\label{sub3sub1sec2}
Referring to Table \ref{tab_bench_nlp}, for the AG's News Corpus and Reuters Corpus Volume 1 under the labeling crude vs all, the $M_2^{(\lambda, \sigma^2)}$ model family, particularly $M_2^{(0.5, 2.0)}$ is preferred. These parameter selections suggest a slightly more precision centric penalty. When considering Reuters Corpus Volume 1 with the labeling crude vs all and the Stanford Sentiment Treebank, there is no observed improvement. In the case of the Hate Speech Data, a more distinctive context, there is roughly a 4\% boost under the $M_2^{(2.0, 2.0)}$ model. This parameter selection is also a balanced penalty between recall and precision. Overall, similar to the Image benchmark conclusion, the $M_2^{(5.0, 5.0)}$ is a frequent performer in the extra variation set of parameters. This insight of balanced penalty selection also holds for contextual text data.

\begin{sidewaystable}
\sidewaystablefn%
\begin{center}
\begin{minipage}{\textwidth}
\caption{Best F1 Score for $M_1^\beta$ and $M_2^{(\lambda, \sigma^2)}$ over $30$ Epochs}\label{tab_bench_nlp} 
\begin{tabular*}{\textwidth}{@{\extracolsep{\fill}}lcccccccccc@{\extracolsep{\fill}} }
\toprule
&\multicolumn{1}{@{}c@{}}{Baseline} & \multicolumn{5}{@{}c@{}}{Parameter Variations Section \ref{sub3sec5}} & \multicolumn{4}{@{}c@{}}{Extra Variations}  \\ \cmidrule{2-2}\cmidrule{3-7}\cmidrule{8-11}
Dataset\footnotemark[0]      & $M_B$     & $M_1^{16}$ & $M_2^{(0.5, 0.5)}$ & $M_2^{(0.5, 2.0)}$ & $M_2^{(2.0, 0.5)}$ & $M_2^{(2.0, 2.0)}$     & $M_1^{8}$ & $M_1^{32}$ &  $M_2^{(0.01, 0.01)}$ &  $M_2^{(5.0, 5.0)}$ \\
\midrule
ag\_news\footnotemark[1] &0.9632 &0.9474 &0.9632 &\textbf{0.9639} &0.9626 &0.9624 &0.9553 &0.9404 &0.9634 &\textbf{0.9655} \\
rcv1\footnotemark[2] &0.9333 &0.9298 &0.9396 &\textbf{0.9461} &0.9211 &0.9451 &0.9316 &0.927 &0.9356 &\textbf{0.9501} \\
rcv1\footnotemark[2] &\textbf{0.9324} &0.92 &0.9251 &0.9189 &0.9178 &0.9189 &0.9127 &0.9139 &0.9251 &0.9054 \\
hate\footnotemark[3] &0.8671 &0.8304 &0.8741 &0.9045 &0.8621 &\textbf{0.9046} &0.8383 &0.7669 &0.8655 &\textbf{0.8868} \\
sst\footnotemark[4] &\textbf{0.8175} &0.7619 &0.7955 &0.7909 &0.8071 &0.8001 &0.7727 &0.7494 &0.8018 &0.8004 \\
\botrule
\end{tabular*}
\footnotetext[0] {Datasets are found in Hugging Face repository. The \textbf{base-url} is \url{https://huggingface.co/datasets}. }
\footnotetext[1]{Train/test 90K/30K, label 1 25\%, labeling is 3 vs all, AG's News Corpus Data found here \textbf{base-url}/ag\_news.  }
\footnotetext[2]{Train/test 5485/2189 \& 5485/2189, label 1 4.61\% \& 4.57\%, labeling is crude vs all \& trade vs all, Reuters Corpus Volume 1 Data found here \textbf{base-url}/yangwang825/reuters-21578. }
\footnotetext[3]{Train/test 8027/2676, label 1 11\%, labeling is 1 vs 0, Hate Speech Data found here \textbf{base-url}/hate\_speech18. }
\footnotetext[4]{Train/test 67K/872, label 1 55\%, labeling is 1 vs 0, Stanford Sentiment Treebank found here \textbf{base-url}/sst2. }
\end{minipage}
\end{center}
\end{sidewaystable}

\subsubsection{Structured/Tabular Results}\label{sub3sub1sec3}
The tabular or structured benchmark results in Table \ref{tab_bench_tab} show that this paper's methodology outperforms the baseline for all but one dataset (the breast cancer dataset). A key insight is that, for the parameter variations from section \ref{sub3sec5} and the extra variations, a more recall centric penalty is preferred. In particular, the $M_1^\beta$ and $M_2^{(2.0, 0.5)}$ model families for the datasets iono, pima, vehicle, glass, vowel, yeast and abalone are favored. The remaining datasets - seg and sat - show modest improvement for the balanced penalty or $M_2^{(0.5, 2.0)}$ model. Compared to the Census results in Table \ref{tab2}, it appears that feature distinctiveness plays a major part for tabular data. This paper defines feature distinctiveness as a neural network learning better discriminative features with respect to the dependent variable. This conclusion arises from the more recall centric penalty showing up in the result, suggesting that for tabular or structured data, the network should focus on learning strong discriminative features to enhance recall. This result underscores the hypothesis of this paper that the type of data, particularly contextual data, matters for a metric-based penalty and further supports the flexibility of this $F_\beta$ penalty methodology.

\begin{sidewaystable}
\sidewaystablefn%
\begin{center}
\begin{minipage}{\textwidth}
\caption{Best F1 Score for $M_1^\beta$ and $M_2^{(\lambda, \sigma^2)}$ over $30$ Epochs}\label{tab_bench_tab} 
\begin{tabular*}{\textwidth}{@{\extracolsep{\fill}}lcccccccccc@{\extracolsep{\fill}} }
\toprule
&\multicolumn{1}{@{}c@{}}{Baseline} & \multicolumn{5}{@{}c@{}}{Parameter Variations Section \ref{sub3sec5}} & \multicolumn{4}{@{}c@{}}{Extra Variations}  \\ \cmidrule{2-2}\cmidrule{3-7}\cmidrule{8-11}
Dataset\footnotemark[0]      & $M_B$     & $M_1^{16}$ & $M_2^{(0.5, 0.5)}$ & $M_2^{(0.5, 2.0)}$ & $M_2^{(2.0, 0.5)}$ & $M_2^{(2.0, 2.0)}$     & $M_1^{8}$ & $M_1^{32}$ &  $M_2^{(0.01, 0.01)}$ &  $M_2^{(5.0, 5.0)}$ \\
\midrule
iono\footnotemark[1]  &0.7845 &\textbf{0.8364} &0.8068 &0.8161 &0.8092 &0.8256 &0.8205 &\textbf{0.8742} &0.8114 &0.7845 \\
pima\footnotemark[2]  &0.5253 &0.4407 &0.4109 &0.3645 &\textbf{0.5454} &0.5088 &0.5124 &0.2711 &0.5058 &0.5208 \\
breast\footnotemark[3]  &\textbf{0.9416} &0.7985 &0.6464 &0.8633 &0.7934 &0.9387 &0.7832 &0.8239 &0.7589 &0.7832 \\
vehicle\footnotemark[4]  &0.3942 &0.4423 &0.4000 &0.3363 &\textbf{0.4507} &0.3470 &0.2105 &0.3247 &0.3103 &0.3333 \\
seg\footnotemark[5] &0.6798 &0.5099 &0.6078 &\textbf{0.6987} &0.5571 &0.5295 &0.3915 &0.3130 &0.6645 &0.3247 \\
glass\footnotemark[6] &0.8695 &0.7200 &0.7407 &0.7826 &\textbf{0.9473} &0.6250 &0.9473 &0.7000 &0.8333 &\textbf{0.9523} \\
sat\footnotemark[7] &0.5511 &0.1674 &0.3274 &\textbf{0.5571} &0.4963 &0.1313 &0.2375 &0.1714 &\textbf{0.5849} &0.3779 \\
vowel\footnotemark[8] &0.2752 &\textbf{0.3439} &0.3076 &0.2926 &0.3103 &0.2434 &0.2464 &0.1851 &0.1647 &\textbf{0.2979} \\
yeast\footnotemark[9] &0.5491 &\textbf{0.8717} &0.7500 &0.5079 &0.2185 &0.2010 &\textbf{0.6046} &0.5084 &0.6857 &0.2105 \\
abalone\footnotemark[10] &0.9723 &0.9723 &0.9723 &0.9723 &0.9723 &0.9723 &0.9723 &\textbf{0.9765} &0.9723 &0.9723 \\
\botrule
\end{tabular*}
\footnotetext[0]{Data urls: \textbf{UCI-url} \url{https://archive.ics.uci.edu/datasets/} or \textbf{R-url} \url{https://github.com/cran/mlbench/tree/master/data/}.}
\footnotetext[1]{Train/test 235/116, label 1 34\%, Ionosphere Data found in \textbf{UCI-url}.}
\footnotetext[2]{Train/test 514/254, label 1 35\%, Pima Indians Diabetes Data found in \textbf{R-url}. }
\footnotetext[3]{Train/test 381/188, label 1 38\%, Breast Cancer Wisconsin Data found in \textbf{UCI-url}.}
\footnotetext[4]{Train/test 566/280, label 1 27\%, labeling is opel vs all, Vehicle Data found in \textbf{R-url}. }
\footnotetext[5]{Train/test 210/2100, label 1 14\%, labeling is brickface vs all, Segmentation Data found in \textbf{UCI-url}.}
\footnotetext[6]{Train/test 143/71, label 1 13\%, labeling is 7 vs all, Glass Data found in \textbf{R-url}. }
\footnotetext[7]{Train/test 4308/1004, label 1 9\%, labeling is 4 vs all, Satellite Data found in \textbf{UCI-url}.}
\footnotetext[8]{Train/test 663/327, label 1 9\%, labeling is hYd vs all, Vowel Data found in \textbf{R-url}.}
\footnotetext[9]{Train/test 344/170 label 1 9\%, labeling is CYT vs ME2, Yeast Data found in \textbf{UCI-url}.}
\footnotetext[10]{Train/test 489/242, label 1 6\%, labeling is 18 vs 9, Abalone Data found in \textbf{UCI-url}.}
\end{minipage}
\end{center}
\end{sidewaystable}

\newpage

\section{Conclusion}\label{sec7}
This paper proposes a weighted cross-entropy based on van Rijsbergen $F_\beta$ measure. By assuming statistical distributions as an intermediary, an optimal $\beta$ can be found, which is then used as a penalty weighting in the loss function. This approach is convenient since van Rijsbergen defines $\beta$ to be a weighting parameter between recall and precision. Guided training by the $F_\beta$ hypothesizes that the interaction of the many combinations between the minority and majority classes has information that can help in three ways. First, as in \citeauthor{bib105} it can improve feature selection. Second, model training can generalize better. Lastly, overall performance may improve. Results from Table \ref{tab2} show that this methodology helps in achieving better $F_1$ scores in some cases, with the added benefit of parameter interpretation from $M_1^\beta$ and $M_2^{(\lambda, \sigma^2)}$. Furthermore, when considering results from real-life use cases as in Table \ref{tab_bench_real}, commonalities between model families start to surface. Parameter selections that yield recall-centric penalties for both $M_1^\beta$ and $M_2^{(\lambda, \sigma^2)}$ can be observed. The analyses from this paper provide the following insights: (1) the balanced penalty distribution is a good starting point for $M_2^{(\lambda, \sigma^2)}$ model family, (2) feature distinctiveness impacts parameter selections between both model families, (3) non-contextual data such as tabular or structured data, seem to benefit from a recall centric penalty, (4) $M_1^\beta$ may be better for image data, and $M_2^{(\lambda, \sigma^2)}$ for text, and (5) contextual-based data are better positioned for embedding architectures than non-contextual data - except when the tabular data can be mapped to contextual data or the features are discriminative.These points show that $F_\beta$ as a performance metric can be integrated alongside a loss function through penalty weights by using statistical distributions.
\newpage

\bibliography{sn-bibliography}

\newpage

\begin{appendices}

\section{General form of F-Beta: \textit{n-th} derivative}\label{secA1}
The derivation pattern using Sasaki's \cite{bib2} steps are straightforward for any partial derivative after the first derivative. To set the stage, a few equations are listed. 
\begin{itemize}
\item[--] From (\ref{eq1}), it can easily be shown that $\frac{1}{\alpha \frac{1}{p} + (1-\alpha) \frac{1}{r}}=\frac{pr}{\alpha r+ (1-\alpha) p}$. 
\item[--] Keeping the notation similar to \cite{bib2}, let $g = \alpha r+ (1-\alpha) p$ then $\frac{\partial g}{\partial r} = \alpha$ and $\frac{\partial g}{\partial p} = 1-\alpha$.
\item[--] Taking the first derivative of  (\ref{eq1}) via the chain rule yields the following: $\frac{\partial E}{\partial r} = \frac{-pg + pr\frac{\partial g}{\partial r}}{g^2}$ and $\frac{\partial E}{\partial p} = \frac{-pg + pr\frac{\partial g}{\partial p}}{g^2}$.
\item[--] After simplifying, $\frac{\partial E}{\partial r} = \frac{-(1-\alpha)p^2}{g^2}$ and $\frac{\partial E}{\partial p} = \frac{-\alpha r^2}{g^2}$.
\end{itemize}
After setting $\frac{\partial^n E}{\partial r^n}$=$\frac{\partial^n E}{\partial p^n}$ for $n=1$ it's easy to see that $(1-\alpha)p^2=\alpha r^2$ and using $\beta=\frac{r}{p}$ yields $\alpha$ that pertains to the original $F_\beta$ measure or (\ref{eq2}) with $n=1$. With the same steps, for $n=2$ the equality becomes $2(1-\alpha)p^2 \alpha = 2\alpha r^2 (1-\alpha)$ or $p=r$ implying $\beta=1$. With each successive differentiation where $n \gt 2$, the pattern is as follows: $c \alpha^{n-2}p^2 = c (1-\alpha)^{n-2}r^2$, where $c$ is the same constant on both sides. Using $r = \beta p$ will then give the generalized equality $\alpha_n = \frac{1}{\beta^\frac{-2}{n-2} + 1}$.

\section{Case 1: Joint Probability Distribution for U and IU}\label{secA2}
To prove (\ref{eq6a}) it is sufficient to set up both integrals and explain the bounds. The computation itself is straightforward. From the following probability $\text{Pr}\left( F_\beta \right) = \text{Pr}\left( F_\beta \leq z \right) = \text{Pr}\left( X_1 X_2 \leq z \right)$ it is clear that the domain is in $[\frac{r^\prime}{r+\beta^*}, \frac{r^\prime+\beta^*}{r}]$ based on (\ref{eq4}) and (\ref{eq5}). With a slight rearrangement, we can say the following: 
\begin{align}
\text{Pr}\left( F_\beta \right) &= \text{Pr}\left( X_1 X_2 \leq z \right) \nonumber \\
&=  \int_{r^\prime}^{r^\prime + \beta^*} \text{Pr}\left( x_2 \leq \frac{z}{x_1} \right) f_{x1} dx_1, \nonumber
\end{align}
where $f_{x_1}$ is the probability density of $X_1$ and $\text{Pr}\left( x_2 \leq \frac{z}{x_1} \right)$ will be the cumulative distribution for $X_2$. These are quite common and can be found online, or in \cite{dud} or \cite{hogg}. The bounds come from (\ref{eq4}). Using these bounds, notice that for $x_2 \leq \frac{z}{x_1}$ to exist, then $\frac{z}{x_1} \geq \frac{1}{r+\beta^*}$ and  $\frac{z}{x_1} \leq \frac{1}{r}$. This results in the range $rz \leq x_1 \leq (r+\beta^*)z$. Recall, that the existence of $x_1$ is in the range $r^\prime \leq x_1 \leq r^\prime + \beta^*$. From both intervals on $x_1$, define condition 1 as $rz \leq r^\prime$ or $z \leq p$ and condition 2 as $(r+\beta^*)z \leq r^\prime + \beta^*$ or $\frac{(r+\beta^*)z}{r^\prime + \beta^*} \leq 1$. We need to consider separately the following scenarios: condition 1 and 2 are true, condition 1 and 2 are both false, and condition 1 is false and condition 2 is true. The scenario of condition 1 being true and condition 2 being false does not occur.

\begin{proof}[Proof]
For $z \leq p$ \& $\frac{(r+\beta^*)z}{r^\prime + \beta^*} \leq 1$ we get the following: 
\begin{align}
&\int_{r^\prime}^{r^\prime+\beta^*} \text{Pr}\left( x_2 \leq \frac{z}{x_1} \right) f_{x1} dx_1 = \dfrac{1}{\beta^*} \int_{r^\prime}^{(r+\beta^*)z} \text{Pr}\left( x_2 \leq \frac{z}{x_1} \right) dx_1, \nonumber \\
&= \dfrac{1}{(\beta^*)^2} \int_{r^\prime}^{(r+\beta^*)z} \left( \left( r+\beta^* \right) - \frac{x_1}{z}  \right) dx_1 \nonumber \\
&= \dfrac{ \frac{z}{2} \left(r + \beta^* - \frac{r^\prime}{z} \right)^2}{(\beta^*)^2}. \nonumber
\end{align}

For $z \gt p$ \& $\frac{(r+\beta^*)z}{r^\prime + \beta^*} \gt 1$ we get the following: 
\begin{align}
&\int_{r^\prime}^{r^\prime+\beta^*} \text{Pr}\left( x_2 \leq \frac{z}{x_1} \right) f_{x1} dx_1 \nonumber \\
&= \dfrac{1}{\beta^*} \int_{r^\prime}^{rz} dx_1 + \dfrac{1}{\beta^*} \int_{rz}^{r^\prime+\beta^*} \text{Pr}\left( x_2 \leq \frac{z}{x_1} \right) dx_1, \nonumber \\
&= \left( \dfrac{rz - r^\prime}{\beta^*} + \dfrac{1}{\left( \beta^*\right)^2} \left(    \left(  r+ \beta^* + \frac{r^\prime + \beta^*}{2z}  \right) \left( r^\prime + \beta^* - rz \right)   \right. \right. \nonumber \\
&+ \left. \left. \dfrac{r \left( rz - \left( r^\prime + \beta^* \right) \right)}{2}  \right) \right) \nonumber
\end{align}

For $z \gt p$ \& $\frac{(r+\beta^*)z}{r^\prime + \beta^*} \leq 1$ we get the following: 
\begin{align}
&\int_{r^\prime}^{r^\prime+\beta^*} \text{Pr}\left( x_2 \leq \frac{z}{x_1} \right) f_{x1} dx_1 \nonumber \\
&= \dfrac{1}{\beta^*} \int_{r^\prime}^{rz} dx_1 + \dfrac{1}{\beta^*} \int_{rz}^{r^\prime+\beta^*} \text{Pr}\left( x_2 \leq \frac{z}{x_1} \right) dx_1 - \dfrac{1}{\beta^*} \int_{(r+\beta^*)z}^{r^\prime+\beta^*} \text{Pr}\left( x_2 \leq \frac{z}{x_1} \right) dx_1, \nonumber \\
&= \dfrac{rz - r^\prime}{\beta^*} + \dfrac{1}{\left( \beta^*\right)^2} \left(    \left(  r+ \beta^* + \frac{r^\prime + \beta^*}{2z}  \right) \left( r^\prime + \beta^* - rz \right)   \right. \nonumber \\
&+ \left. \dfrac{r \left( rz - \left( r^\prime + \beta^* \right) \right)}{2}  \right) - \dfrac{1}{\left( \beta^*\right)^2} \left( (r+\beta^*)(r^\prime +\beta^*) - \frac{(r^\prime + \beta^*)^2}{2z} - \frac{(r+\beta^*)^2 z}{2} \right) \nonumber
\end{align}

For the scenario $z \leq p$ \& $\frac{(r+\beta^*)z}{r^\prime + \beta^*} \gt 1$, we need to show that it never occurs. By rearranging condition 2, and recalling $r^\prime = pr$, we can get $r(z-p) > \beta^*(1-z)$. Then, if $z \leq 1$ then $r(z-p) \leq 0$ by $z \leq p$. Since $\beta^* \gt 0$, $r(z-p) > \beta^*(1-z)$ never occurs. If, $z \gt 1$ then it implies $p \gt 1$ since $z \leq p$. This never occurs since $p \in [0,1]$.
\end{proof}

\newpage
\section{Case 2: Joint Probability Distribution for Ga and IE}\label{secA3}
The derivation of (\ref{eq8}) is similar to Case 1 in that the integral will be broken into pieces and probability distribution proof will be used again. As before using the same rearrangement, we can say the following: 
\begin{align}
\text{Pr}\left( F_\beta \right) &= \text{Pr}\left( X_1 X_2 \leq z \right) \nonumber \\
&=  \int_{-\infty}^{+\infty} \text{Pr}\left( x_2 \leq \frac{z}{x_1} \right) f_{x1} dx_1, \nonumber
\end{align}
Before moving forward, $X_2$'s marginal distribution or (\ref{eq7}) will be given. 
\begin{align}
&\text{If, } \beta^{\prime\prime} \sim \text{Exponential}(\lambda), \text{ then the CDF for } X=\beta^{\prime\prime} + r \text{ is: } \nonumber \\
&F(x) = \text{Pr}(X \leq x) = \text{Pr}(\beta^{\prime\prime} + r \leq x^\prime_2) = \text{Pr}(\beta^{\prime\prime} \leq x-r )  \nonumber \\
&F(x) = 1-\exp \left( - \lambda ( x -r )\right), \forall x \geq r.  \nonumber \\
&\text{With transformation: } Y = g(X) = \frac{1}{X} \text{ so, } g^{-1}(y) = \beta^{\prime\prime} = \frac{1-ry}{y} \nonumber \\
&\text{Then, } F_y(y) = F_x(g^{-1}(y)) = \exp \left(  - \lambda \left( \frac{1-ry}{y} \right) \right) \text{ since $g$ is a} \nonumber \\
&\text{strictly decreasing function.} \nonumber
\end{align}
Now, we can see that $X_2$ has the distribution $F(X_2) = \exp \left(  - \lambda \left( \frac{1-rx_2}{x_2} \right) \right)$ where $x_2$ $\in$ [0,$\frac{1}{r}$]. Using this property we complete the proof.
\begin{proof}[Proof]
\textbf{For  z  $\gt$ 0: }
\begin{align}
&\int_{-\infty}^{rz} f_{x_1} d_{x_1} +  \int_{rz}^{+\infty} \exp \left \{  - \lambda \left( \frac{1-r \frac{z}{x_1}}{\frac{z}{x_1}} \right) \right \} f_{x_1} d_{x_1} \nonumber \\
&= \Phi(rz; r^\prime, \sigma^2) +  \dfrac{1}{\sqrt{2 \pi \sigma^2}} \int_{rz}^{+\infty} \exp \left \{  - \lambda \left( \frac{1-r \frac{z}{x_1}}{\frac{z}{x_1}} \right) \right \} \exp \left \{ -\frac{1}{2 \sigma^2} \left( x_1 -r^\prime \right)^2 \right \}  \nonumber \\
&= \Phi(rz; r^\prime, \sigma^2) +  \dfrac{1}{\sqrt{2 \pi \sigma^2}} \int_{rz}^{+\infty} \exp \left \{ - \frac{x_1^2 - 2 r^\prime x_1 + (r^\prime)^2}{2 \sigma^2} - \frac{\lambda x_1}{z}  +\lambda r   \right \} dx_1 \nonumber \\
&= \Phi(rz; r^\prime, \sigma^2) +  \dfrac{1}{\sqrt{2 \pi \sigma^2}} \int_{rz}^{+\infty} \exp \left \{ - \frac{\left( x_1 - \left( r^\prime - \frac{\lambda \sigma^2}{z} \right) \right)^2  }{2 \sigma^2}  \right. \nonumber \\
& \left. + \frac{\frac{2 r^\prime \lambda \sigma^2}{z} - \left( \frac{\lambda \sigma^2}{z} \right)^2 + \lambda r}{2 \sigma^2} \right \} dx_1 \nonumber \\
&=\left[ \Phi(rz; r^\prime, \sigma^2) + \exp \left( \lambda r + \dfrac{\left( \frac{\lambda \sigma^2}{z} \right)^2 - \frac{2 r^\prime \lambda \sigma^2}{z}}{2 \sigma^2} \right) \right. \nonumber \\
& \left. \times  \left( 1 - \Phi \left( rz; \left( r^\prime - \frac{\lambda \sigma^2}{z} \right), \sigma^2  \right)   \right) \right]. \nonumber
\end{align}
To be clear, the bounds of the integral around $rz$ arise from the directionally based inequality on $\frac{z}{x_1}$, in particular $\frac{z}{x_1} \gt \frac{1}{r}$. \\ \\
\textbf{For  z=0: } it can be seen that the entire mass is summarized by the Gaussian distribution or $X_1$ since $X_2$ is non-negative.
\begin{align}
\int_{-\infty}^{0} f_{x_1} d_{x_1} = \Phi(0; r^\prime, \sigma^2). \nonumber
\end{align}
\textbf{For  z $\lt$ 0: } this is a bit different because though the Inverse Exponential redistributes the mass for the Gaussian as before, it being non-negative needs to be adjusted for. Consider the interval $[-\infty, 0]$ that represents the domain for this (probability) mass. Let $p_z$ be the probability mass of interest. Next, define $b_1$ to be the mass from $[-\infty, rz]$, $b_2$ to be the mass from $[rz, 0]$, and $b_3$ as the mass for $[-\infty, 0]$. Notice for $b_1$ and $b_2$, the separation of the integral is similar as before but with different bounds. So we have the following:
\begin{align}
&b_1 = \int_{-\infty}^{rz} \exp \left \{  - \lambda \left( \frac{1-r \frac{z}{x_1}}{\frac{z}{x_1}} \right) \right \} f_{x_1} d_{x_1} \nonumber \\
&b_2 = \int_{rz}^{0} f_{x_1} d_{x_1} \nonumber \\
&b_3 = \Phi(0; r^\prime, \sigma^2). \nonumber
\end{align}
By $X_2$'s redistribution, the mass for the negative values is $p_z = b_3 - b_1 - b_2$ for $z \lt 0$. The proof is now simplified to solving the $p_z$ expression and by using some of the results from the $z \gt 0$ case we have the following: 
\begin{align}
&\Phi(0; r^\prime, \sigma^2) -  \int_{-\infty}^{rz} \exp \left \{  - \lambda \left( \frac{1-r \frac{z}{x_1}}{\frac{z}{x_1}} \right) \right \} f_{x_1} d_{x_1}  - \int_{rz}^{0} f_{x_1} d_{x_1}\nonumber \\
&= \Phi(0; r^\prime, \sigma^2) - \exp \left( \lambda r + \dfrac{\left( \frac{\lambda \sigma^2}{z} \right)^2 - \frac{2 r^\prime \lambda \sigma^2}{z}}{2 \sigma^2} \right)  \times \Phi \left( rz; \left( r^\prime - \frac{\lambda \sigma^2}{z} \right), \sigma^2  \right)  \nonumber \\
&- \left( \Phi \left(0; r^\prime, \sigma^2 \right) - \Phi \left( rz;  r^\prime, \sigma^2\right) \right) \nonumber \\
&=\Phi(rz; r^\prime, \sigma^2) - \exp \left( \lambda r + \dfrac{\left( \frac{\lambda \sigma^2}{z} \right)^2 - \frac{2 r^\prime \lambda \sigma^2}{z}}{2 \sigma^2} \right) \times \Phi \left( rz; \left( r^\prime - \frac{\lambda \sigma^2}{z} \right), \sigma^2  \right)  \nonumber
\end{align}
\end{proof}

\newpage
\section{Pressure Vessel Design}\label{secA4}
This is borrowed from \cite{bib103}. See figure 7 from \citeauthor{bib103} to see the structure design of pressure vessel, which looks similar to Underground Storage Tanks discussed earlier. 
\subsection{Problem Statement}\label{secA4_1}
The pressure vessel design objective is to minimize total cost, which includes material, forming and welding. The design variables are thickness of the shell ($T_s$), thickness of the head ($T_h$), the inner radius ($R$) and the length of the cylinder ($L$). The mathematical formulation is found in \ref{pveq}.
\begin{align}
X &= [x_1, x_2, x_3, x_4] = [T_s, T_h, R, L]  \notag \\
f(X) &= 0.6224x_1x_3x_4 + 1.7781x_2x_3^2 + 3.1661x_1^2x_4 + 19.84x_1^2x_3 \label{pveq}
\end{align}
For this paper, \citeauthor{bib103} HAOASMA algorithm results will be used as the baseline for the parameters. These parameter results serve as the best minimum result. To be specific, $T_s = 1.8048$, $T_h = 0.0939$, $R = 13.8360$, and $L = 123.2019$. The next section will provide a couple variations to convert this problem to a classification problem.

\subsection{Varying Design Parameter Plots}\label{secA4_1}
\begin{figure}[H]
\centering
\includegraphics[width=0.35\textwidth]{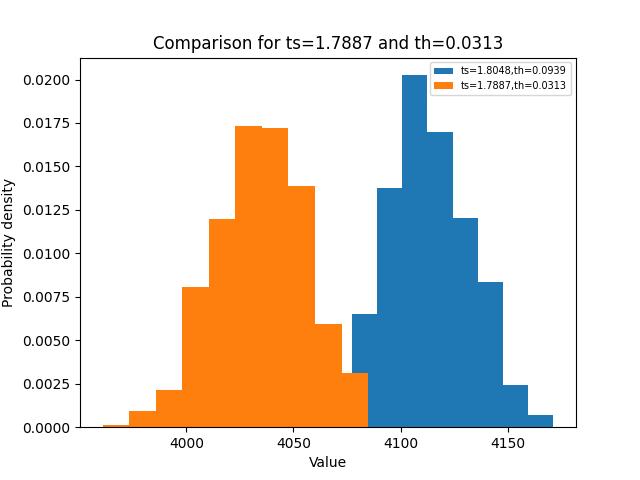 }
\includegraphics[width=0.35\textwidth]{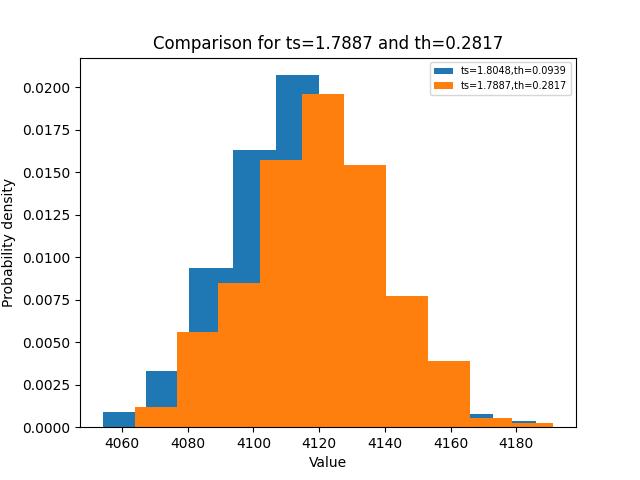 }
\caption{Comparison of a simulated realization using Algorithm \ref{algo2} and variations to the thickness parameter. Left: $T_s^v=1.7887$ and $T_h^v=0.0313$ and Right: $T_s^v=1.7887$ and $T_h^v=0.2817$ }
\label{pv}
\end{figure}

The simulation carried out can be done using Algorithm \ref{algo2}. Two simulated realizations from this algorithm can be seen in Figure \ref{pv}. The left figure has values $T_s^v=1.7887$ and $T_h^v=0.0313$ and the right figure has $T_s^v=1.7887$ and $T_h^v=0.2817$. The superscript $v$ stands for variation. The variations are intended to be reflect two scenarios: the first is a clear separation between distribution (the left figure), hence an easier classification. The second has significant overlap (the right figure) or a tougher classification.

\begin{algorithm}
\caption{Simulation of Pressure Vessel Data for Classification}\label{algo2}
\begin{algorithmic}[1]
\Require $s \gt 0$ $\wedge$ $i \in(0,1)$ and $T_s^v > 0$ and $T_h^v > 0$ where $i$ is imbalance, $s$ is the size of the data set and $T_s^v$ and $T_h^v$ are parameter variations from the HAOASMA baseline.
\State Set $T_s^b = 1.8048$, $T_h^b = 0.0939$, $R^b = 13.8360$, and $L^b = 123.2019$ where superscript $b$ is baseline.
\State Compute data label sizes based on the imbalance $i$ as $s_0 = \lfloor s\times(1-i) \rfloor$ and $s_1 = s-s_0$ where the subscript $0$ and $1$ stand for data sizes for label 0 and label 1.
\State Initialize $t_s^b$ and $t_h^b$ as $s_0$ size arrays with values $T_s^b$ and $T_h^b$ respectively. 
\State Draw $r^b$ and $l^b$ arrays of size $s_0$ from the normal distributions: $N(\mu=R^b, \sigma^2=1)$ and $N(\mu=L^b, \sigma^2=1)$ respectively.
\State Concatenate the column vectors $t_s^b$, $t_h^b$, $r^b$ and $l^b$ together and assign this array to variable $X_0$.
\State Apply \ref{pveq} to each row of $X_0$ to yield a cost value and assign this array to variable $Y_0$.
\State Initialize $l_0$ as a label 0 array of size $s_0$ with the value $0$.
\State Concatenate the column vectors $r^b$, $l^b$, $Y_0$, and $l_0$ together and assign this array to variable $F_0$.
\State Initialize $t_s^v$ and $t_h^v$ as $s_1$ size arrays with values $T_s^v$ and $T_h^v$ respectively. 
\State Draw $r^b$ and $l^b$ arrays of size $s_1$ from the normal distributions: $N(\mu=R^b, \sigma^2=1)$ and $N(\mu=L^b, \sigma^2=1)$ respectively.
\State Concatenate the column vectors $t_s^v$, $t_h^v$, $r^b$ and $l^b$ together and assign this array to variable $X_1$.
\State Apply \ref{pveq} to each row of $X_1$ to yield a cost value and assign this array to variable $Y_1$.
\State Initialize $l_1$ as a label 1 array of size $s_1$ with the value $1$.
\State Concatenate the column vectors $r^b$, $l^b$, $Y_1$, and $l_1$ together and assign this array to variable $F_1$.
\State Concatenate or stack the arrays $F_0$ and $F_1$ which yields a total of $s$ rows with the last column being the labels for classification.
\end{algorithmic}
\end{algorithm}

\cleardoublepage

\section{Underground Storage Tank (UST)}\label{secA5}
This is section is borrowed from \cite{ramdh1} and all equations and derivations and further explorations can be found there.

\subsection{Problem Statement}\label{secA5_1}
The UST problem deals with estimating tank dimensions by using only vertical height measurements. It is also possible that this cylindrical UST has hemispherical endcaps appended on the ends which will also contain volume. The equations for the volume based on cross-sectional measurements for the tanks with the cylindrical, cylindrical with hemispherical endcaps, ellipsoidal, and ellipsoidal with hemi-ellipsoidal endcaps shapes, are given in (\ref{fc}), (\ref{fch}), (\ref{fed}), and (\ref{fedh}), respectively. \\ 

The equation for the Cylindrical shape is:
\begin{align}
&f_{C}(r,L,h)= L \left \{r^2 cos^{-1} \left(\frac{r-h}{r} \right) -(r-h)\sqrt{2rh-h^2} \right \}   \label{fc}
\end{align}

If one were to add hemispherical endcaps to the cylinder ends the subsequent volume would be:
\begin{align}
f_{CH}(r,L,h)=  L \left \{r^2 cos^{-1} \left(\frac{r-h}{r} \right) -(r-h)\sqrt{2rh-h^2} \right \} + \frac{\pi h^2}{3}(3r-h) \label{fch} 
\end{align}

The equation for the Elliptical shape for a deformed Cylinder is: 
\begin{align}
f_{ED}(a,b,L,h)&=  L \left \{(ab) cos^{-1} \left(\frac{a-h}{\sqrt{a^2+ (h^2 -2ha) \left( 1- \frac{b^2}{a^2}\right)} } \right) \right. \nonumber \\
&\left. -b(a-h)\sqrt{1- \left( 1-\frac{h}{a}\right)^2} \right \}   \label{fed}
\end{align}

If one were to add hemispherical endcaps to the cylinder which deforms to hemi-ellipsoidal endcaps the subsequent volume would be:
\begin{align}
f_{EDH}(a,b,L,h)&=  L \left \{(ab) cos^{-1} \left(\frac{a-h}{\sqrt{a^2+ (h^2 -2ha) \left( 1- \frac{b^2}{a^2}\right)} } \right)  \right. \nonumber \\
&\left. -b(a-h)\sqrt{1- \left( 1-\frac{h}{a}\right)^2} \right \} + \dfrac{2 \pi a^3 + \pi (a-h) \left( \frac{h b^2}{a} \right) \left( \frac{h-2a}{a} \right) }{3} \notag \\
&- \dfrac{2 \pi a^3 (a-h)}{3\sqrt{a^2+ (h^2 -2ha) \left( 1- \frac{b^2}{a^2}\right)}} \label{fedh}
\end{align}

\subsection{Varying Tank Dimension}\label{secA5_2}
The parameters used in the Algorithm \ref{algo3} are borrowed from \cite{ramdh1} and the baseline will be the cylindrical case with radius, $r=4$ and length, $L=32$ and the parameter variations will be on the $a$ and $b$ for an ellipse. \citeauthor{ramdh1} used a measurement error based model for simulation, which will also be used here. The measurement errors will be on the heights $h$. Similar to the pressure vessel we consider an easy and a tough simulation scenario for classification. This is seen in Figure \ref{cyl} where the left figure is easier to distinguish between cylinder and ellipse versus the right figure. The same interpretation can be seen for the end-cap based equations or Figure \ref{cylend}. 

\begin{algorithm}
\caption{Simulation of Tank Dimension Data for Classification}\label{algo3}
\begin{algorithmic}[1]
\Require $s \gt 0$ $\wedge$ $i \in(0,1)$ and $a > 0$ and $b > 0$ where $i$ is imbalance, $s$ is the size of the data set and $a$ and $b$ are the parameters for the vertical and horizontal axis of an ellipse.
\State Set $r=4$, $L=32$ as the baseline parameters. 
\State Compute data label sizes based on imbalance $i$ as $s_0 = \lfloor s\times(1-i) \rfloor$ and $s_1 = s-s_0$ where the subscript $0$ and $1$ stand for data sizes for label 0 and label 1.
\State Initialize noise arrays $\epsilon_0 \sim N(0, 2)$ and $\gamma_0 \sim U(-0.05, 0.05)$ both of size $s_0$.
\State Draw an array of heights $h_0 \sim U(1, 2\times r-1)$ of size $s_0$ for the vertical height of a cylinder.
\State Compute variable $h_0^\prime = h_0 + \gamma_0$.
\State Initialize $l_0$ as a label 0 array of size $s_0$ with the value $0$.
\State Compute the volume from either (\ref{fc}) or (\ref{fch}) using $h_0$, $r$, $L$ and assign this to variable $X_0$.
\State Assign variable $Y_0 = X_0 + \epsilon_0$.
\State Concatenate the column arrays $Y_0$, $h_0^\prime$, and $l_0$ and assign this to $F_0$.
\State Initialize noise arrays $\epsilon_1 \sim N(0, 2)$ and $\gamma_1 \sim U(-0.05, 0.05)$ both of size $s_1$.
\State Draw an array of heights $h_1 \sim U(1, 2\times a-1)$ of size $s_1$ for the vertical height of an ellipse.
\State Compute variable $h_1^\prime = h_1 + \gamma_1$.
\State Initialize $l_1$ as a label 1 array of size $s_1$ with the value $1$.
\State Compute the volume from either (\ref{fed}) or (\ref{fedh}) using $h_1$, $a$, $b$, $L$ and assign this to variable $X_1$.
\State Assign variable $Y_1 = X_1 + \epsilon_1$.
\State Concatenate the column arrays $Y_1$, $h_1^\prime$, and $l_1$ and assign this to $F_1$.
\State Concatenate or stack the arrays $F_0$ and $F_1$ which yields a total of $s$ rows with the last column being the labels for classification.
\end{algorithmic}
\end{algorithm}

\begin{figure}[H]
\centering
\includegraphics[width=0.30\textwidth]{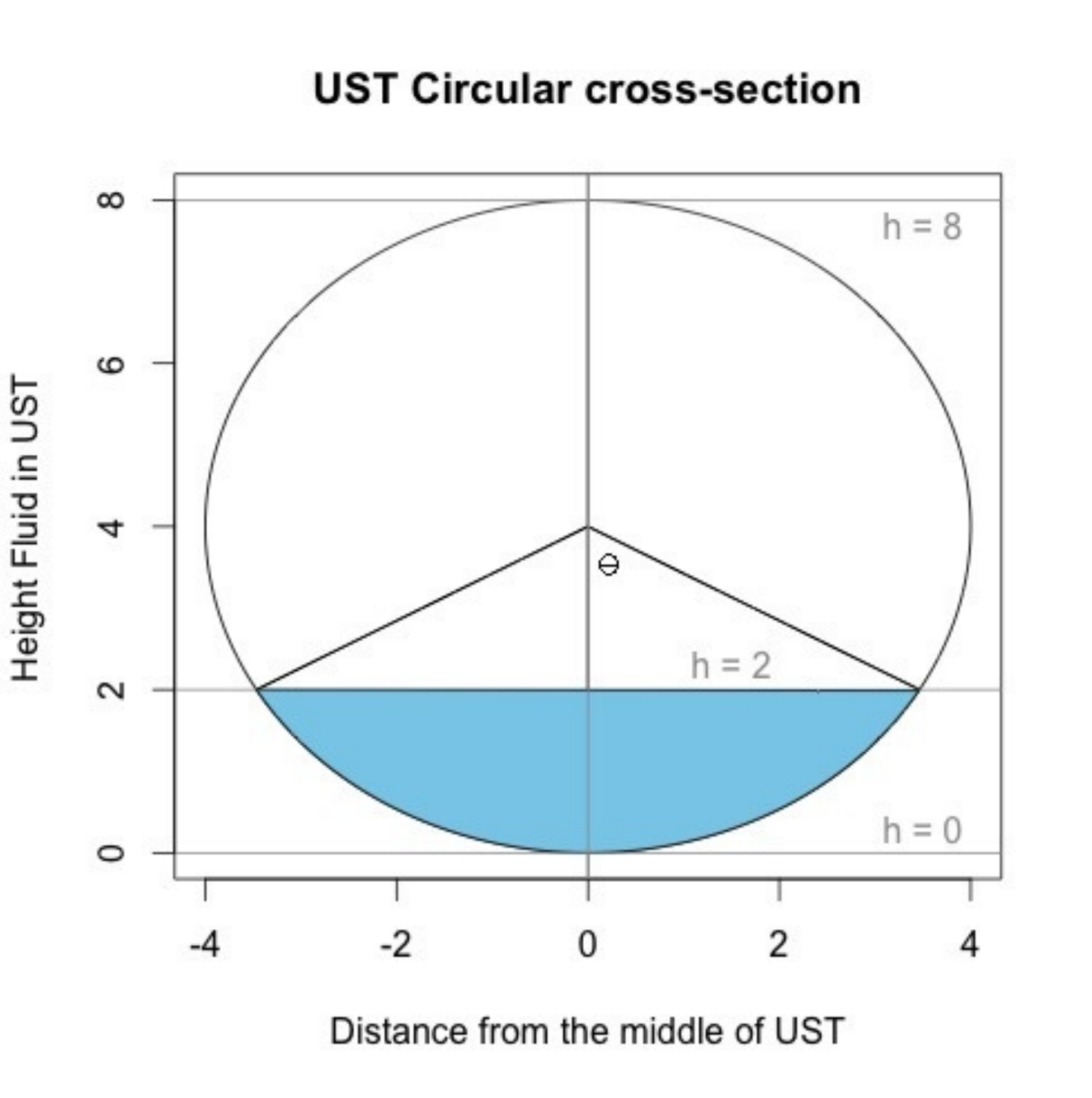 }
\includegraphics[width=0.41\textwidth]{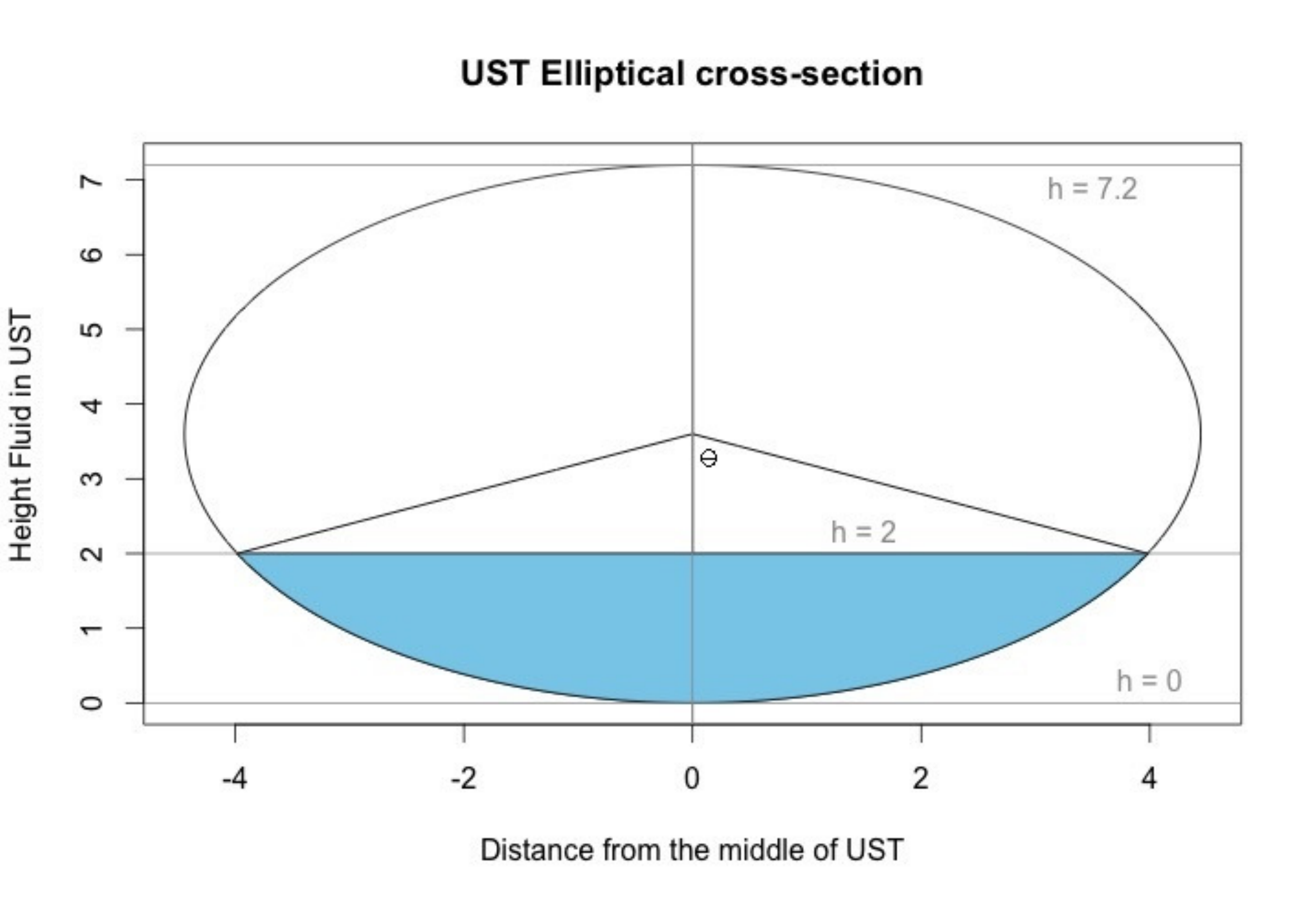 }
\caption{Left: cylindrical UST with a radius (r) of 4 and total max height of 8. Right: elliptical UST (deformed) with vertical axis (a) of 3.6 and max height of 7.2. }
\label{histogram}
\end{figure}

\begin{figure}[H]
\centering
\includegraphics[width=0.35\textwidth]{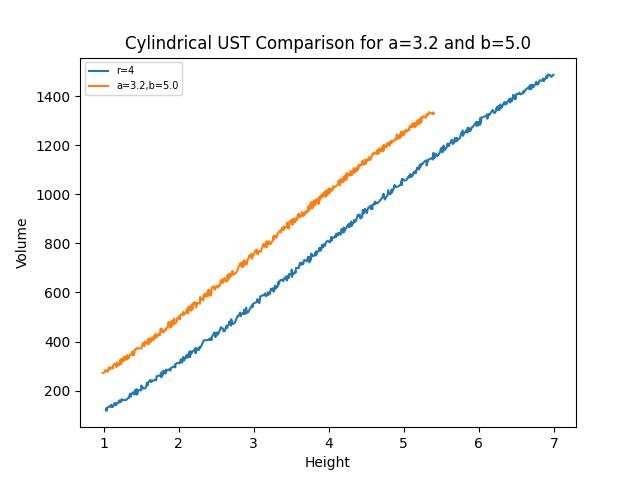 }
\includegraphics[width=0.35\textwidth]{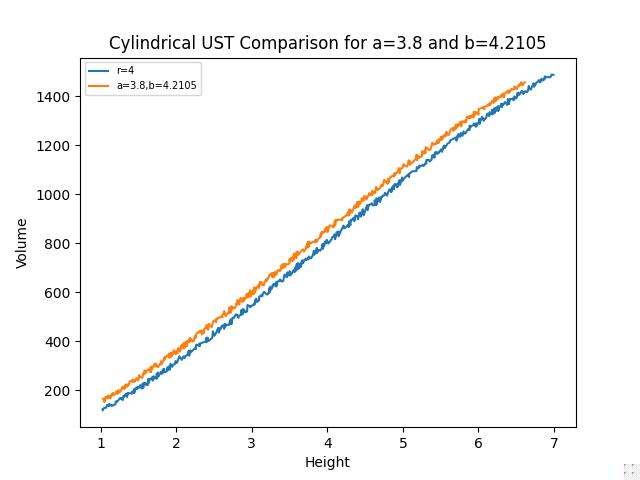 }
\caption{Comparison of one simulated realization of volume versus height measurements using Algorithm \ref{algo3} for equation (\ref{fc}) versus (\ref{fed}). Left: $a=3.2$, $b=5.0$ versus $r=4$ and Right: $a=3.8$, $b=4.2105$ versus $r=4$.}
\label{cyl}
\end{figure}

\begin{figure}[H]
\centering
\includegraphics[width=0.35\textwidth]{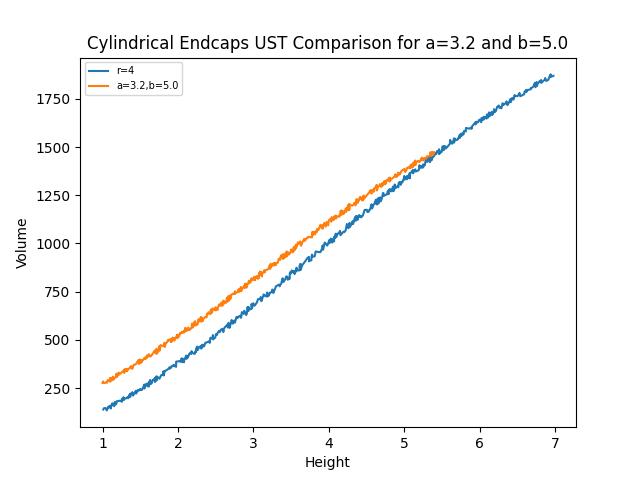 }
\includegraphics[width=0.35\textwidth]{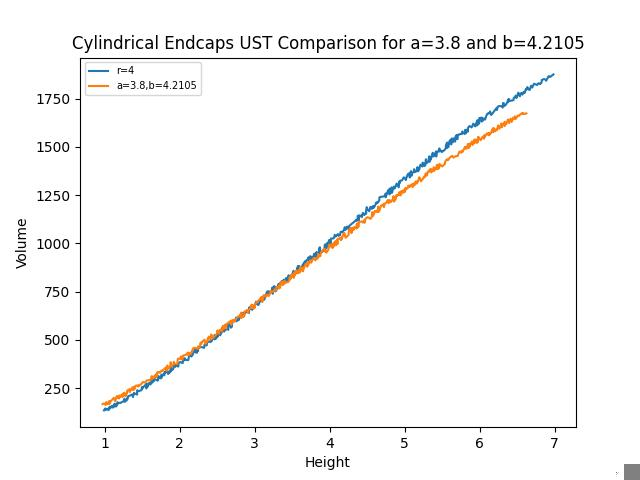 }
\caption{Comparison of one simulated realization of volume versus height measurements using Algorithm \ref{algo3} for equation (\ref{fch}) versus (\ref{fedh}). Left: $a=3.2$, $b=5.0$ versus $r=4$ and Right: $a=3.8$, $b=4.2105$ versus $r=4$.}
\label{cylend}
\end{figure}




\end{appendices}



\end{document}